%% file: arxiv.tex
\documentclass[10pt,twocolumn,letterpaper]{article}

\usepackage[pagenumbers]{wacv} %

\usepackage{graphicx}
\usepackage{amsmath}
\usepackage{amssymb}
\usepackage{booktabs}\usepackage{times}
\usepackage{epsfig}
\usepackage{colortbl}
\usepackage{amsthm,amsfonts}
\usepackage{enumitem}
\usepackage{makecell, rotating, multirow, diagbox}
\usepackage[ruled, vlined, linesnumbered]{algorithm2e}%
\newtheorem{theorem}{Theorem}

\newtheorem{proposition}{Proposition}

\newtheorem{remark}{Remark}
\usepackage{paralist}
\usepackage[dvipsnames]{xcolor}
\usepackage{longtable}

\usepackage[pagebackref=true,breaklinks=true,letterpaper=true,colorlinks,bookmarks=false]{hyperref}
\definecolor{yancolor1}{rgb}{0.87, 0.43, 0.26}
\definecolor{yancolor2}{rgb}{0.31, 0.43, 0.48}

\usepackage{bm}
\usepackage{dsfont}
\newcommand{\ind}{\mathds{1}}
\usepackage{microtype}

\usepackage[numbers,sort&compress]{natbib} 

\usepackage[capitalize]{cleveref}
\crefname{section}{Sec.}{Secs.}
\Crefname{section}{Section}{Sections}
\Crefname{table}{Table}{Tables}
\crefname{table}{Tab.}{Tabs.}

\def\wacvPaperID{71} %
\def\confName{WACV}
\def\confYear{2024}

\begin{document}

\title{Improving Fairness in Deepfake Detection\thanks{\textcolor{blue}{This paper has been accepted by WACV 2024}}}

\author{
Yan Ju$^{1}$\thanks{ Equal contribution} , \ Shu Hu$^{2\dagger \ddagger}$, \ Shan Jia$^{1}$, \ George H. Chen$^{3}$, \ Siwei Lyu$^{1}$\thanks{Corresponding authors}\\
$^1$ University at Buffalo, State University of New York {\tt \small \{yanju, shanjia, siweilyu\}@buffalo.edu}\\
$^2$ Indiana University–Purdue University Indianapolis {\tt \small hu968@purdue.edu}\\
$^3$ Carnegie Mellon University {\tt \small georgechen@cmu.edu} \\
}
\maketitle

\begin{abstract} 
Despite the development of effective deepfake detectors in recent years, recent studies have demonstrated that biases in the data used to train these detectors can lead to disparities in detection accuracy across different races and genders. This can result in different groups being unfairly targeted or excluded from detection, allowing undetected deepfakes to manipulate public opinion and erode trust in a deepfake detection model. While existing studies have focused on evaluating fairness of deepfake detectors, to the best of our knowledge, no method has been developed to encourage fairness in deepfake detection at the algorithm level. In this work, we make the first attempt to improve deepfake detection fairness by proposing novel loss functions that handle both the setting where demographic information (\eg, annotations of race and gender) is available as well as the case where this information is absent. Fundamentally, both approaches can be used to convert many existing deepfake detectors into ones that encourages fairness. Extensive experiments on four deepfake datasets and five deepfake detectors demonstrate the effectiveness and flexibility of our approach in improving deepfake detection fairness. Our code is available at \url{https://github.com/littlejuyan/DF_Fairness}.
\end{abstract}

\setlength{\abovedisplayskip}{2.5pt plus 1.5pt}
\setlength{\belowdisplayskip}{2.5pt plus 1.5pt}
\setlength{\abovedisplayshortskip}{1.5pt plus 1pt}
\setlength{\belowdisplayshortskip}{1.5pt plus 1pt}

\vspace{-3mm}
\section{Introduction}\label{sec:intro}
\vspace{-1mm}

``Deepfakes" refer to realistic images and videos where a person's likeness has been replaced by that of another with the help of deep learning technologies. %
Concerns have arisen regarding deepfakes being used for malicious purposes, such as in political propaganda or cyberattacks. For example, a deepfake video can depict a world leader making statements or taking actions that never occurred in reality~\cite{bohavcek2022protecting}, which could deceive the public. To mitigate the impact of deepfakes, a variety of deepfake detectors have been developed with promising detection accuracy~\cite{masood2022deepfakes, nguyen2019use, coccomini2022combining, wang2022m2tr, zhao2021multi,li2020sharp, pu2022learning, guo2022open, guo2022eyes, guo2022robust, wang2022gan, hu2021exposing}. %

However, recent studies \cite{trinh2021examination, xu2022comprehensive, nadimpalli2022gbdf,masood2022deepfakes, zhang2023x, yang2023improving} have shown that current deepfake detectors are unfair: their detection accuracy is not consistent across gender, age, and ethnicity~\cite{xu2022comprehensive}. For example, several state-of-the-art detectors have higher detection accuracy for deepfakes with lighter skin tones than deepfakes with darker skin tones~\cite{hazirbas2021towards, trinh2021examination}.   
A key reason for this disparity is that how often different demographic groups appear in the training data is imbalanced~\cite{nadimpalli2022gbdf}. Collecting a larger ``balanced'' dataset can be costly and labor-intensive~\cite{news1}. While conventional fairness methods %
can be applied (\eg, by adding a fairness regularization term to the overall loss function \cite{mehrabi2021survey}), deepfake detection poses an additional level of complexity. Specifically, we need to account for the imbalance in real vs training deepfake examples in addition to the usual imbalance in demographic groups.

In this paper, we propose two Fair Deepfake Detection (FDD) methods, both of which can be used to modify an existing deep-learning-based deepfake detector that does not account for fairness into one that does:
\begin{compactenum}
\item Our first method DAG-FDD (\underline{d}emographic-\underline{ag}nostic FDD) does not rely on demographic details (the user does not have to specify which attributes to treat as sensitive such as race and gender) and can be applied when, for instance, these demographic details have not been collected for the dataset. %
To use DAG-FDD, the user needs to specify a probability threshold for a minority group without explicitly identifying all possible groups. The goal is to ensure that all groups with at least a specified occurrence probability have low error. %

\item The second method DAW-FDD (\underline{d}emographic-\underline{aw}are FDD) leverages demographic information and employs an existing fairness risk measure~\cite{williamson2019fairness}. At a high level, DAW-FDD aims to ensure that the losses achieved by different user-specified groups of interest (\eg, different races or genders) are similar to each other (so that the deepfake detector is not more accurate on one group vs another) and, moreover, that the losses across all groups are low. This approach requires a way to estimate the loss of each group, for which we use a ranking-based estimator that addresses the imbalance in real vs deepfake examples per group.

\end{compactenum}
From a technical viewpoint, both of our methods are based on a distributionally robust optimization (DRO) technique called \textit{Conditional Value-at-Risk} (CVaR) \cite{rahimian2019distributionally,levy2020large, rockafellar2000optimization}. Whereas our first method DAG-FDD is a straightforward application of CVaR to the fair deepfake detection problem (so that the novelty is not in the method itself but in applying the method to a problem that we do not believe has previously been explored by DRO literature), our second method DAW-FDD uses the CVaR in a hierarchical manner that, to the best of our knowledge, is novel. Specifically, DAW-FDD uses a CVAR loss function across groups (to address imbalance in demographic groups) and, per group, DAW-FDD uses another CVAR loss function (to address imbalance in real vs deepfake training examples). We also show how several existing fairness approaches are special cases of DAW-FDD.

Our main contributions are as follows:
\begin{compactenum}
\item We propose two methods for achieving fair deepfake detection in ways that are either agnostic to or, separately, aware of demographic factors. Both methods convert an existing deep-learning-based deepfake detector that does not encourage fairness into one that does. Moreover, both use training procedures that alternate between minibatch gradient descent (to update neural network model parameters) and solving specific convex optimization problems related to data imbalance. %
\item We demonstrate the effectiveness of our methods in improving fairness of several state-of-the-art deepfake detectors (while retaining strong detection performance) on four large-scale datasets (FaceForensics++~\cite{rossler2019faceforensics++}, Celeb-DF~\cite{li2020celeb}, DeepFakeDetection~(DFD)~\cite{dfd}, and Deepfake Detection Challenge (DFDC)~\cite{dolhansky2020deepfake}). %

\end{compactenum}
To the best of our knowledge, our paper is the first to propose novel algorithms for fair deepfake detection.

\vspace{-2mm}
\section{Related Work}\label{sec:relatedworks}
\vspace{-1mm}
\subsection{The Categories of Fairness Approaches }
\vspace{-1mm}

Many approaches have been proposed to encourage fairness in general machine learning settings. These methods fall into two major categories: demographic-agnostic and demographic-aware. Typically, there is a tradeoff between encouraging fairness and achieving high prediction accuracy.

\smallskip
\vspace{-1mm}\noindent
\textbf{Demographic-agnostic}. When demographic information is inaccessible (\eg, either it was not collected, or we do not have an exhaustive list of all groups that we want to be ``fair'' across), there are methods that achieve fairness without any prior knowledge of which attributes to treat as sensitive. %
Examples of such approaches include distributionally robust optimization (DRO) \cite{hashimoto2018fairness}, adversarial learning \cite{lahoti2020fairness}, using input features to find surrogate group information \cite{zhao2022towards}, cluster-based balancing for input data \cite{yan2020fair}, knowledge distillation \cite{chaifairness}, and causal variational autoencoders \cite{grari2021fairness}. Among these, our work builds on existing DRO literature. %

\smallskip
\vspace{-1mm}\noindent
\textbf{Demographic-aware}. A large number of fairness definitions have been proposed in the literature for generating regularization terms to add to a training loss. There are two key types of fairness measures using demographic information that we highlight: group fairness \cite{dwork2012fairness} and intersectional fairness \cite{foulds2020intersectional}. Specifically, group fairness considers a model fair across a user-specified set of groups if these different groups satisfy a condition such as demographic parity \cite{dwork2012fairness} or equalized odds \cite{hardt2016equality}. Intersectional fairness accounts for multiple sensitive attributes (\eg, intersections of race and gender taking on specific combinations). %
More notions of fairness can be found in \cite{mehrabi2021survey, caton2020fairness}. A drawback of these approaches is that precisely which notion of fairness and which attributes to treat as sensitive (or an exhaustive list of groups to encourage fairness across) must be specified as part of the overall training loss function. If at a later time, we realize that we want to use a different notion of fairness or we want to account for different sensitive attributes or demographic groups, then model re-training may be required. One of our proposed approaches requires an exhaustive list of all groups that we want similar accuracy for.

\vspace{-1mm}
\subsection{Fairness in Deepfake Detection}
\vspace{-1.5mm}

Despite considerable efforts~\cite{masood2023deepfakes,dong2022protecting,zhao2021multi,dong2023implicit,wang2023altfreezing} dedicated to enhancing the generalization capability of deepfake detection to out-of-distribution (OOD) data, there is still limited progress in addressing the biased performance during testing within known domains. In contrast to previous studies, our research uniquely prioritizes fairness as its primary goal. Specifically, we address the issue of biased performance among groups under in-domain testing, aiming to achieve equal accuracy across user-specified demographic groups.

\begin{table}[t]
\renewcommand\arraystretch{1}
\center
\newcommand{\tabincell}[2]{\begin{tabular}{@{}#1@{}}#2\end{tabular}} 
\scalebox{0.6}{
\begin{tabular}{ c | c | c  |c | c|c }
 \hline
{Method} & {Year} & \tabincell{c}{\#Detector} & \tabincell{c}{\#Dataset} &  \makecell{Fairness\\Solution}& \makecell{Require\\Demographics}\\
\hline
  Trinh~\textit{et al}.~\cite{trinh2021examination} & 2021 & 3 & 1 &   $\times$ & -\\

Hazirbas~\textit{et al}.~\cite{hazirbas2021towards} & 2021 & 5 & 1  & $\times$ &  - \\

Pu~\textit{et al}.~\cite{pu2022fairness} & 2022 & 1 & 1  &  $\times$ &  - \\

GBDF~\cite{nadimpalli2022gbdf} & 2022 & 5 & 4  & Data-level&   \checkmark  \\

 Xu~\textit{et al}.~\cite{xu2022comprehensive} & 2022 & 3 & 4 &  $\times$&  -\\
\hline
\rowcolor{gray!20} \tabincell{c}{\textbf{DAG-FDD} (ours)} & {2023} & {5} & {4} &  {Algorithm-level} &  $\times$ \\

\rowcolor{gray!20} \tabincell{c}{\textbf{DAW-FDD} (ours)} & 2023 & 5 & 4 & Algorithm-level & \checkmark \\
\hline
\end{tabular}
}
\vspace{-0.3cm}
\caption{\label{tb:relatedworks} \textit{Summary of previous studies and our work. `-' means not applicable.}}
\vspace{-7mm}
\label{tab: t1}
\end{table}

Extending the investigation of fairness that was originally for face recognition~\cite{yu2020fair, karkkainen2021fairface, fang2022fairness, ramachandra2022algorithmic}, several recent studies have examined fairness concerns in deepfake detection, as shown in Table \ref{tab: t1}. The work in~\cite{trinh2021examination} is the first to evaluate biases in existing deepfake datasets and detection models across protected groups. 
They examined three popular deepfake detectors and observed large disparities in prediction accuracy across races, with up to 10.7\% difference in error rate between groups. Similar observations are found in~\cite{hazirbas2021towards}.
Pu \etal \cite{pu2022fairness} evaluated the reliability of one popular deepfake detection model (MesoInception-4) on FF++ and showed that the MesoInception-4 model is generally more effective for female subjects. A more comprehensive analysis of deepfake detection bias with regards to demographic and non-demographic attributes is presented in \cite{xu2022comprehensive}. The authors collected comprehensive annotations for 5 widely-used deepfake detection datasets to facilitate future research.
The work in~\cite{nadimpalli2022gbdf} showed significant bias in both datasets and detection models and they tried to reduce the performance bias across genders by providing a gender-balanced dataset. This leads to limited improvement at the cost of highly time-consuming data annotation, which does not extend to other possible non-gender attributes that we might want to treat as sensitive. Developing more effective bias-mitigating deepfakes detection solutions remains an open challenge~\cite{masood2022deepfakes}.

\vspace{-1mm}
\section{Method}\label{sec:method}
\vspace{-2mm}

In this work, we propose two deep-learning-based deepfake detection methods that encourage fairness. %
The first method, termed DAG-FDD, is applicable when we have training data without demographic annotations. This approach works with most existing deepfake datasets. The second method, termed DAW-FDD, works when the dataset contains additional demographic annotations (specifically so that we know which group each training point belongs to among some user-specified exhaustive list of groups we aim to ensure fairness over).

Both approaches are meant to modify an existing deep-learning-based deepfake detector into one that encourages fairness. To this end, in what follows, we assume that $\mathcal{S} := \{(X_i,Y_i)\}_{i=1}^n$ is the training set that consists of i.i.d.~samples from a joint distribution $\mathbb{P}$, where $X_i$ is the $i$-th data point's raw features (\eg, an image or video) and $Y_i\in\{0,1\}$ is the \mbox{$i$-th} point's label (0 means real, 1 means deepfake). We assume that the underlying deepfake detector aims to minimize a risk of the form
\begin{equation}
\mathcal{R}_{\text{avg}}(\theta)
:= \mathbb{E}_{(X,Y)\sim\mathbb{P}}[\ell(\theta;X,Y)]
\quad\text{for }\theta\in\Theta,
\label{eq:avg-risk}
\end{equation}
where $\ell$ is the loss function (\eg, cross entropy loss) of the deepfake detector model, which is assumed to have parameters $\theta$ that belong to a set $\Theta$; the loss function is evaluated for a specific input~$X$ with target label~$Y$. As is standard in machine learning, instead of minimizing the true unknown risk $\mathcal{R}_{\text{avg}}(\theta)$, in practice we use some variant of minibatch gradient descent to minimize the empirical risk given by the loss function $\mathcal{L}_{\text{avg}}(\theta) := \frac{1}{n}\sum_{i=1}^n \ell(\theta, X_i, Y_i)$.

\vspace{-1mm}
\subsection{Demographic-agnostic FDD (DAG-FDD)}\label{sec:FDD-without-demographics}
\vspace{-2mm}

We first present DAG-FDD, which is based on the distributionally robust optimization (DRO) \cite{hashimoto2018fairness, duchi2021learning}. Roughly, the idea is that there are $K$ unknown underlying groups of individuals. We assume that each group occurs with probability at least $\alpha\in(0,1)$. Then by a standard result of DRO, there is a loss function that we can minimize that aims to ensure that all $K$ latent groups have low error despite us not explicitly knowing what these latent groups are. We formalize this high level idea in the rest of this section.

\smallskip
\noindent
\textbf{The worst-case risk }$\mathcal{R}_{\max}(\theta)$\textbf{.} 
We assume that there are $K$ true unknown groups that comprise the joint distribution~$\mathbb{P}$. In other words,~$\mathbb{P}$ can be represented as a mixture of $K$ distributions $\mathbb{P}:=\sum_{m=1}^K\pi_m \mathbb{P}_m$,
where the $m$-th group occurs with probability $\pi_m\in(0,1)$ and has distribution~$\mathbb{P}_m$, and $\sum_{m=1}^K\pi_m=1$. Instead of minimizing the average risk~\eqref{eq:avg-risk}, we seek to minimize the following worst-case risk:
\begin{equation}
    \begin{aligned}
    \mathcal{R}_{\max}(\theta):= \max_{m=1,...,K}\mathbb{E}_{(X,Y)\sim \mathbb{P}_m}[\ell(\theta; X,Y)],
    \end{aligned}
\label{eq:worst-case risk}
\end{equation} 
where $\ell$ is a loss function for an individual data point used by the original deepfake detector that we are modifying (\ie, $\ell$ is same function used in equation \eqref{eq:avg-risk}). Directly minimizing $\mathcal{R}_{\max}$ is intractable since we do not know the $K$ latent groups; in fact, we assume that we do not know the value of $K$ either. However, it turns out that we can minimize an empirical version of its upper bound.

\smallskip
\noindent
\textbf{Upper bound on }$\mathcal{R}_{\max}(\theta)$\textbf{.}
We use the well-established risk function called the \textit{Conditional Value-at-Risk} (CVaR)~\cite{rockafellar2000optimization}:
\begin{equation}
\text{CVaR}_{\alpha}(\theta):=\inf_{\lambda\in \mathbb{R}}\Big\{\lambda+\frac{1}{\alpha}\mathbb{E}_{(X,Y)\sim\mathbb{P}}\big[[\ell(\theta;X,Y)-\lambda]_+\big]\Big\},
\label{eq:cvar}
\end{equation}
where $[a]_+=\max\{0,a\}$ is the hinge function (also called the ReLU function), and we assume that each of the $K$ latent groups occurs with probability at least $\alpha\in(0,1)$. %
The following result shows that the risk CVaR$_{\alpha}(\theta)$ is an upper bound of $\mathcal{R}_{\max}(\theta)$, so that by minimizing CVaR$_{\alpha}(\theta)$, we are minimizing an upper bound on the worst-case risk \eqref{eq:worst-case risk}.

\vspace{-2mm}
\begin{proposition}
Suppose that $\alpha\leq \min_{m=1,...,K}\pi_m$. Then $\emph{CVaR}_{\alpha}(\theta) \geq \mathcal{R}_{\max}(\theta)$.
\label{prop:cvar-risk}
\end{proposition} \vspace{-2mm}\noindent
Note that this result is not new \cite{zhai2021doro}. However, to the best of our knowledge, we are the first to apply it to learning fair deepfake detectors in a demographic-agnostic way. We include the proof of Proposition \ref{prop:cvar-risk} in Appendix \ref{appendix:prop cvar-risk}.

In practice, %
$\alpha$ is a user-specified hyperparameter that says how rare of a group we want to ensure low risk for. As $\alpha\rightarrow0$, we are asking for low risk even for an extremely rare group. In contrast, as $\alpha\rightarrow1$ (\ie, the rarest group occurs with probability~1), then for Proposition~\ref{prop:cvar-risk} to hold, it means that we would have $K=1$ and $\pi_1=1$, in which case the worst-case risk~\eqref{eq:worst-case risk} would simply become the standard average risk~\eqref{eq:avg-risk}. By tuning $\alpha\in(0,1)$, we effectively say how ``fine-grain'' of groups we aim to encourage fairness over, which naturally leads to a tradeoff between fairness and population-level average accuracy.

\begin{algorithm}[t]\footnotesize
    \caption{DAG-FDD}\label{alg:DAG-FDD}
    \SetAlgoLined
    \KwIn{A training dataset $\mathcal{S}$ of size $n$, $\alpha$, max\_iterations, num\_batch, learning rate $\eta$}
    \KwOut{A fair deepfake detection model with parameters $\theta^*$} 
    
    \textbf{Initialization:} $\theta_0$, $l=0$

    \For{$e=1$ to \emph{max\_iterations}}{
    \For{$b=1$ to \emph{num\_batch}}{
    { \mbox{Sample a mini-batch $\mathcal{S}_b$ from $\mathcal{S}$}}
    
    Compute $\ell(\theta_l;X_i,Y_i)$, $\forall (X_i,Y_i)\in \mathcal{S}_b$
    
    Using binary search to find $\lambda$ that minimizes (\ref{eq:cvar_empirical}) on $\mathcal{S}_b$
    
    Compute $\theta_{l+1}$ with equation (\ref{eq:update-rule-DAG-FDD})
    
    $l\leftarrow l+1$
    }
    }
    \Return{$\theta^* \leftarrow \theta_{l}$}
\end{algorithm}

\smallskip
\noindent
\textbf{DAG-FDD.} In practice, we minimize an empirical version of CVaR$_{\alpha}(\theta)$. This gives us the following optimization problem, which we refer to as our first method DAG-FDD:
\begin{equation}
\min_{\theta\in\Theta, \lambda\in \mathbb{R}} \mathcal{L}_{\text{DAG-FDD}}(\theta,\lambda)\!:=\!\lambda\!+\!\frac{1}{\alpha n}\sum_{i=1}^n[\ell(\theta;X_i,Y_i)-\lambda]_+.
\label{eq:cvar_empirical}
\end{equation}
As a reminder, $\Theta$ is the set of possible model parameters of the original deepfake detector (see equation~\eqref{eq:avg-risk}).

To provide some intuition for the loss function $\mathcal{L}_{\text{DAG-FDD}}(\theta,\lambda)$, suppose for a moment that we have obtained the optimal value of $\lambda^*$ in \eqref{eq:cvar_empirical}. Then the only training points that contribute to the loss are the ``hard'' ones with a loss value greater than $\lambda^*$. %
In other words, the loss function always focuses on ``hard'' training points with large enough loss values whereas the ``easy'' training points with low loss values are ignored. Which training points are ``easy'' vs ``hard'' can vary as a function of model parameters $\theta\in\Theta$.

Solving the optimization problem in  (\ref{eq:cvar_empirical}) can be done through an iterative gradient descent approach \cite{hu2021tkml, hu2022distributionally, hu2022sum, hu2020learning}.
In practice, we first initialize model parameters $\theta$ and then randomly select a mini-batch set $\mathcal{S}_b$ from the training set $\mathcal{S}$, performing the following two steps for each iteration on $\mathcal{S}_b$ (see Algorithm \ref{alg:DAG-FDD}):
\begin{compactitem}
\item We fix $\theta$ and use binary search to find the global optimum of $\lambda$ since $\mathcal{L}_{\text{DAG-FDD}}(\theta,\lambda)$ is convex w.r.t. $\lambda$.
\item We fix $\lambda$ and update $\theta$ using (stochastic) gradient descent with a user-specified learning rate $\eta>0$:
\begin{equation}
\theta_{l+1}=\theta_{l}-\frac{\eta}{\alpha |\mathcal{S}_b|}\sum_{i\in \mathcal{S}_b} \partial_{\theta}\ell(\theta_{l};X_i, Y_i)\cdot \ind_{[\ell(\theta_{l};X_i, Y_i)>\lambda]},
\label{eq:update-rule-DAG-FDD}
\end{equation}
where $\ind_{[a]}$ is an indicator function (that equals 1 if $a$ is true, and 0 otherwise), $\partial_{\theta}\ell$ represents the (sub)gradient of $\ell$ w.r.t. $\theta$, and $\eta$ is the learning rate.
\end{compactitem}
We stop iterating after reaching some user-specified stopping criteria (\eg, maximum number of iterations). Note that this optimization process is similar to how one would train the original deepfake detector being modified except that we now have an additional binary search to update $\lambda$; thus the training time complexity is comparable.

\vspace{-1mm}
\subsection{Demographic-aware FDD (DAW-FDD)}
\vspace{-1.5mm}

We now turn to the setting where within the training data, we have demographic labels available so that we know for each training point which group it belongs to among some user-specified exhaustive listing of all possible groups that we want to ensure fairness across. Specifically, we let $\mathcal{G}$ denote the user-specified set of groups (\eg, if we aim to encourage fairness across gender, then $\mathcal{G}$ would consist of the different genders). Then for the $i$-th training point (with raw features $X_i$ and target label $Y_i$) we assume that we also know its group $G_i \in \mathcal{G}$.

We first state the group-level risk that we aim to minimize that encourages fairness across groups (\ie, this risk aims to address imbalance in user-specified demographic groups within the data). When it comes to empirically estimating this risk, we then discuss how we account for imbalance in real vs deepfake examples.

\smallskip
\noindent
\textbf{Group-level risk (addresses demographic imbalance).} Each group $g\in\mathcal{G}$ has a group-specific risk defined as $\mathcal{R}_g(\theta):=\mathbb{E}_{(X,Y)| G=g}[\ell(\theta;X,Y)]$, where random variable~$G$ denotes the group corresponding to a generic data point with raw features $X$ and target $Y$. To treat the different groups to be ``equally weighted'', the risk we use intentionally views $G$ to be sampled uniformly at random from $\mathcal{G}$ (even if in the actual data, $G$ may not be uniformly distributed so that different groups could occur with different probabilities). Specifically, we use the ``group CVaR'' risk
\begin{equation}
\text{CVaR}_\alpha^\mathcal{G}(\theta)
:=\inf_{\lambda\in\mathbb{R}}\Big\{\lambda+\frac{1}{\alpha}\mathbb{E}_{G\sim\text{Uniform}(\mathcal{G})}\big[[\mathcal{R}_G(\theta)-\lambda]_+
\big]\Big\}.
\label{eq:group-cvar}
\end{equation}
To provide some intuition for this risk, recall that the non-group-level CVaR risk from earlier (equation~\eqref{eq:cvar}) focuses on individual data points that have ``large enough'' loss values (specifically, if $\lambda^*$ achieves the infimum value in equation~\eqref{eq:cvar}, then CVaR$_\alpha(\theta)$ only focuses on data points with loss values above~$\lambda^*$). In the group-level version of CVaR presented in equation~\eqref{eq:group-cvar}, we instead focus on groups that have ``large enough'' risk values.

\smallskip
\noindent
\textbf{The group CVaR risk as a fairness risk measure.} In fact, the group CVaR risk can be directly interpreted as a fairness risk measure, as shown by \cite{williamson2019fairness}.

\vspace{-2mm}\begin{proposition} (Equation (21) of \cite{williamson2019fairness})
Let $\alpha\in(0,1)$,
\begin{align*}
&\min_{\theta\in\Theta}
  \emph{CVaR}_\alpha^\mathcal{G}(\theta) \\
&\quad=
\min_{\theta\in\Theta}
  \Big\{
    \mathbb{E}_{G\sim\text{Uniform}(\mathcal{G})}[\mathcal{R}_G(\theta)]
    +
    \mathbb{D}(\{\mathcal{R}_g(\theta) : g\in\mathcal{G}\})
  \Big\},
\end{align*}
where $\mathbb{D}$ is a ``deviation measure'' that looks at how different the different groups' losses are (if they are all the same, then the deviation measure would be 0). Specifically,
\begin{align*}
&\mathbb{D}(\{\mathcal{R}_g(\theta) : g\in\mathcal{G}\}) \\
&\quad:=
\inf_{\lambda\in\mathbb{R}}
  \Big\{
    \lambda + \frac{1}{\alpha|\mathcal{G}|}
    \sum_{g\in\mathcal{G}}[\mathcal{R}_g(\theta) - \overline{\mathcal{R}}(\theta) - \lambda]_+
  \Big\},
\end{align*}
where $\overline{\mathcal{R}}(\theta):=\frac{1}{|\mathcal{G}|}\sum_{g\in\mathcal{G}} \mathcal{R}_g(\theta)$.
\vspace{-2mm}
\end{proposition}

\noindent
This proposition states that minimizing the group-level CVaR risk is equivalent to minimizing a risk that is the sum of two terms: the first term $\mathbb{E}_{G\sim\text{Uniform}(\mathcal{G})}[\mathcal{R}_G(\theta)] = \frac{1}{|\mathcal{G}|}\sum_{g\in\mathcal{G}}\mathcal{R}_g(\theta)$ is the equally weighted average of the different groups' losses, and the second term ${\mathbb{D}(\{\mathcal{R}_g(\theta) : g\in\mathcal{G}\})}$ asks that the different groups' losses are close to each other (\ie, the deepfake detector learned should not be more accurate for one group vs another).

\smallskip
\vspace{-1mm}
\noindent
\textbf{DAW-FDD (empirical estimation of group-level risk that accounts for imbalance in real vs deepfake examples).} Recall that previously when we presented our demographic agnostic approach DAG-FDD, we empirically estimate the CVAR risk from equation~\eqref{eq:cvar} in a straightforward manner with the loss function $\mathcal{L}_{\text{DAG-FDD}}(\theta,\lambda)$ from equation \eqref{eq:cvar_empirical}. Now that we use a group-level CVAR risk instead (given in equation~\eqref{eq:group-cvar}), we have to be more careful with empirically estimating the risk. The issue is that we need an accurate estimate of each group's risk $\mathcal{R}_g(\theta) = \mathbb{E}_{(X,Y)|G=g}[\ell(\theta;X,Y)]$. A naive approach would take an equally weighted average across examples belonging to group~$g$. However, the imbalance in the number of real vs deepfake examples in group~$g$ could bias the estimate of~$\mathcal{R}_g(\theta)$.

To address this issue, we use the average top-$k$ operator \cite{hu2023rank} to estimate group risks instead of the average operator. In more detail, denote the training points in group $g\in\mathcal{G}$ as $\mathcal{I}_g:=\{i=1,\dots,n:G_i = g\}$, the number of points in group $g$ as $n_g:=|\mathcal{I}_g|$, and the set of individual losses in group $g$ as $\ell^{g}(\theta):=\{\ell(\theta;X_i,X_j):i\in\mathcal{I}_g\}$. We further denote the $j$-th largest loss in $\ell^{g}(\theta)$ as $\ell^{g}_{[j]}$ (ties can be broken in any consistent manner). Then we empirically estimate group $g$'s risk $\mathcal{R}_g(\theta)$ with the loss function
\vspace{-2mm}
\begin{equation}
\mathcal{L}_g(\theta)
:= \frac{1}{k_g}\sum_{j=1}^{k_g} \ell^{g}_{[j]}(\theta),
\label{eq:top-k-group-loss}
\vspace{-1mm}
\end{equation}
where $k_g \in \{1,\dots,n_g\}$ is a user-specified integer. This choice of empirical estimate can enhance the influence of the minority class while reducing the influence of the majority class in each group as samples with small loss values are most likely from the majority class per group. Since $k_g$ may vary across groups, we set $k_g=\alpha_g n_g$, where $\alpha_g\in[1/n_g,1]$. In practice, we can set $\alpha_g$ to be the same for all groups and tune it on a predefined grid.

Finally, we solve the following optimization problem which minimizes an empirical estimate of CVaR$_\alpha^{\mathcal{G}}(\theta)$:
\begin{equation}
\min_{\theta\in\Theta,\lambda\in \mathbb{R}}
\mathcal{L}_{\text{DAW-FDD}}(\theta,\lambda)
:=
\lambda+\frac{1}{\alpha |\mathcal{G}|}\sum_{g\in\mathcal{G}}[\mathcal{L}_g(\theta)-\lambda]_+.
\label{eq:DAW-FDD-ori}
\end{equation}
As it turns out, the group specific loss function $\mathcal{L}_g(\theta)$ in 
equation~\eqref{eq:top-k-group-loss} can itself be written as a CVaR loss, as we establish in the following theorem. %

\vspace{-2mm}    
\begin{theorem}\label{theorem:cvar-atk} 
For a set of real numbers $\overline{\ell}=\{\ell_1,...,\ell_q\}$, let $\ell_{[k]}$ denote the $k$-th largest value in $\overline{\ell}$ for $k\in\{1,\dots q\}$. Then we have $\frac{1}{k}\sum_{i=1}^k \ell_{[i]}=\min_{\lambda\in \mathbb{R}} \{\lambda + \frac{1}{k}\sum_{i=1}^q [\ell_i-\lambda]_+\}$. Using this result, optimization problem \eqref{eq:DAW-FDD-ori} is equivalent to

\vspace{-4mm}
\begin{subequations} \small
\begin{align}
&\min_{\theta\in\Theta,\lambda\in \mathbb{R}} \mathcal{L}_{\emph{DAW-FDD}}(\theta,\lambda)\!:=\! \lambda+\frac{1}{\alpha |\mathcal{G}|}\sum_{g\in\mathcal{G}}
      [\mathcal{L}_g(\theta)-\lambda]_+, \label{eq:eq:DAW-FDD1} \\ 
&\emph{s.t.} \ \mathcal{L}_g(\theta)\!\!=\!\! \min_{\lambda_g\in \mathbb{R}}\!\mathcal{L}_g(\theta,\lambda_g)\!:=\!\lambda_g\!+\!\frac{1}{\alpha_g n_g}\!\!\sum_{i\in\mathcal{I}_g}\! [\ell(\theta;X_i,Y_i)-\lambda_g]_+. \label{eq:eq:DAW-FDD2}
\end{align}
\label{eq:eq:DAW-FDD}
\end{subequations}
\end{theorem} \vspace{-4mm} \noindent
We defer the proof to Appendix \ref{proof:theorem:cvar-atk}.
Theorem \ref{theorem:cvar-atk} tells us that optimization problem \eqref{eq:DAW-FDD-ori} is equivalent to a optimization problem with a hierarchical structure: across the demographic groups, we have a group-level CVaR loss (equation~\eqref{eq:eq:DAW-FDD1}). To compute this group-level CVaR loss, we compute each group's loss function $\mathcal{L}_g(\theta)$ (equation~\eqref{eq:eq:DAW-FDD2}), which in turn is of the form of a CVaR loss (\ie, the top-$k$ operator can be written as a CVaR loss). This CVaR loss per group is specifically meant for addressing the imbalance in real vs deepfake examples.
We call this approach DAW-FDD.

To optimize (\ref{eq:eq:DAW-FDD}), the iterative procedure in Algorithm \ref{alg:DAG-FDD} can still be applied except where each iteration now consists of three steps: updating $\{\lambda_g:g\in\mathcal{G}\}, \lambda$, and $\theta$. The pseudocode is shown in Algorithm \ref{alg:DAW-FDD} in Appendix \ref{appendix:algorithms}. Note that the explicit form of $\partial_{\theta}\mathcal{L}_{\text{DAW-FDD}}(\theta_l,\lambda)$ (\ie, the (sub) gradient of $\mathcal{L}_{\text{DAW-FDD}}(\theta,\lambda)$ w.r.t. $\theta$) can be found in Appendix \ref{appendix:explicit_forms}. 

\vspace{-1mm}
\begin{remark}
For our method \emph{DAW-FDD}, by choosing values of $\alpha$ and $\alpha_g$ in specific ways, we recover several existing fairness methods. For example, if $\alpha\rightarrow 1$ and $\alpha_g\rightarrow 1$, we minimize the average of group risks, which aligns with the impartial observer principle \cite{harsanyi1977rational}. If $\alpha\rightarrow0$ and $\alpha_g\rightarrow1$, we instead minimize the largest group risk (\ref{eq:worst-case risk}) \cite{hashimoto2018fairness, mohri2019agnostic}, which aligns with the maximin principle \cite{rawls2001justice}. If $\alpha_g\rightarrow 1$ (\ie, we replace the top-$k$ operator with a simple average), our approach is just the empirical version of $\text{CVaR}_{\alpha}^{\mathcal{G}}(\theta)$ \cite{williamson2019fairness}.
\end{remark}

\vspace{-2mm}
\section{Experiment}
\vspace{-1.5mm}

This section evaluates the effectiveness of the proposed methods in terms of fairness performance and deepfake detection performance. 
We present the most significant information and results of our experiments.
More detailed information and additional results are provided in Appendix \ref{appendix:sec:Additional-Experimental-Details} and \ref{appendix:sec:Additional-Experimental-Results}, respectively. 

\begin{table*}[t]
\renewcommand\arraystretch{1}
\center
\scalebox{0.75}{
\begin{tabular}{c | c | c  c  c | c  c  c | c  c  c | c  c  c  c}
 \hline
 \multirow{3}{*}{Methods} & \multirow{3}{*}{\makecell{Require\\Demo-\\graphics}} & \multicolumn{9}{c|}{Fairness Metrics (\%) \textdownarrow } & \multicolumn{4}{c}{Detection Metrics (\%)} \\ 
 \cline{3-15}
& & \multicolumn{3}{c|}{Gender} & \multicolumn{3}{c|}{Race} & \multicolumn{3}{c|}{Intersection} & \multicolumn{4}{c}{Overall} \\ 
\cline{3-15}
& & $G_{\text{FPR}}$  & $F_{\text{FPR}}$  & $F_{\text{EO}}$  & $G_{\text{FPR}}$  & $F_{\text{FPR}}$  & $F_{\text{EO}}$  & $G_{\text{FPR}}$  &  $F_{\text{FPR}}$  & $F_{\text{EO}}$  & AUC \textuparrow & FPR \textdownarrow & TPR \textuparrow & ACC \textuparrow \\
\hline 
Original & $-$ &  4.10 & 4.10 & 9.06 & 13.09 & 17.28 & 21.00 &  17.93 & 31.59 & 53.95 & 92.76 & 22.06 & 94.43 & 91.49 \\ \hline \hline
$\text{DRO}_{\chi^2}$~\cite{hashimoto2018fairness} & \multirow{2}{*}{$\times$} & 2.68 & 2.68 & 6.75 & 8.32 & 8.97 & 20.40 & 8.73 & 22.97 & 55.54 & 97.18 & 6.32 & 90.25 & 90.86 \\  
 \makecell{DAG-FDD (Ours)} &  &  \colorbox{gray!20}{1.63} & \colorbox{gray!20}{1.63} & \colorbox{gray!20}{6.21}   & \colorbox{gray!20}{8.23} & 9.53 & \colorbox{gray!20}{11.49} & 9.65 & \colorbox{gray!20}{21.21} & \colorbox{gray!20}{48.10} & 97.13 & 9.54 & 94.32  & \colorbox{gray!20}{93.63} \\
\hline \hline
Naive~\cite{nadimpalli2022gbdf}  & \multirow{6}{*}{$\checkmark$} &   11.98 & 11.98 & 18.20 &  16.57 & 22.01 & 25.97 & 28.90 & 72.19 & 93.59 & 83.17 & 50.77 & 92.62 &  84.87 \\
FRM~\cite{williamson2019fairness} & &   1.33 & 1.33 & 5.88 &   9.24 & 12.75 & 20.13 &  10.39 & 25.57 & 60.90 &  \textbf{97.81} &  \textbf{4.76} & 90.85 & 91.63 \\ 
Group DRO~\cite{sagawa2019distributionally} & &   8.20 & 8.20 & 12.87  & 14.37 & 20.97 & 23.79 & 21.86 & 44.98 & 65.24 & 91.13 & 27.83 & 95.15 & 91.04\\
$Cons$. EFPR~\cite{wang2022understanding} &  & 4.24 & 4.24 & 7.91  & 7.09 & 7.49 & 12.41  & 14.95 & 27.80 & 46.62 & 94.30 & 22.61 & 94.94 & 91.80 \\
$Cons$. EO~\cite{wang2022understanding} & &  1.77 & 1.77 & 4.79 & 10.92 & 12.61 & 16.50 & 17.25 & 26.95 & 44.68 & 95.74 & 16.28 & \textbf{95.89} & 93.72 \\
DAW-FDD (Ours) & &  \colorbox{gray!20}{\textbf{0.32}} & \colorbox{gray!20}{ \textbf{0.32}} & \colorbox{gray!20}{ \textbf{3.99}} & \colorbox{gray!20}{ \textbf{2.49}} & \colorbox{gray!20}{ \textbf{3.88}} & \colorbox{gray!20}{ \textbf{6.29}}  & \colorbox{gray!20}{ \textbf{6.61}} & \colorbox{gray!20}{ \textbf{14.06}} & \colorbox{gray!20}{ \textbf{33.84}} & 97.46 & 11.46 & 95.40 & \colorbox{gray!20}{ \textbf{94.17}} \\
 \hline
\end{tabular}
}
\vspace{-0.3cm}
\caption{\label{tb:baselines} \textit{Comparison results with different fairness solutions using Xception detector on FF++ testing set across Gender, Race, and Intersection groups. The best results are shown in \textbf{Bold}. $\uparrow$ means higher is better and $\downarrow$ means lower is better. \colorbox{gray!20}{Gray} highlights mean our methods outperform the baselines in the group (\ie, DAG-FDD vs. Original/$\text{DRO}_{\chi^2}$, DAW-FDD vs. Original/Naive/FRM/Group DRO/$Cons$. EFPR/$Cons$. EO).}} 
\vspace{-6mm}
\end{table*}

\subsection{Experimental Settings}
\vspace{-1mm}

\noindent{\textbf{Datasets.}} Our experiments are based on four popular large-scale benchmark deepfake datasets, namely FaceForensics++ (FF++)~\cite{rossler2019faceforensics++}, Celeb-DF~\cite{li2020celeb}, DeepFakeDetection~(DFD)~\cite{dfd}, and Deepfake Detection Challenge Dataset~(DFDC)~\cite{dolhansky2020deepfake}. Since all the original datasets do not have the demographic information of each video or image, we use the annotations from \cite{xu2022comprehensive} %
which provides annotated demographic information for these four datasets, including Gender (Male and Female) and Race (Asian, White, Black, and Others) attributes. %
{We also double-check the annotations for each dataset}.
In addition to the single attribute fairness, we also consider the combined attributes (Intersection) group, including Male-Asian (M-A), Male-White (M-W), Male-Black (M-B), Male-Others (M-O), Female-Asian (F-A), Female-White (F-W), Female-Black (F-B), and Female-Others (F-O).
We use Dlib~\cite{king2009dlib} for face extraction and alignment, and the cropped faces are resized to $380\times380$ for training and testing. Following the previous study~\cite{xu2022comprehensive}, we split the annotated datasets into training/validation/test sets with a ratio of approximately 60\%/20\%/20\%,  without identity overlapping. In particular, the validation set is used for hyperparameter tuning. More details of the datasets, including attribute groups and number of training samples are provided in Tables~\ref{tb:datasets-details} and \ref{tb:numberofsample} of the Appendix~\ref{appendix:numberoftraining2}.

\noindent{\textbf{Evaluation metrics.}} %
Fairness measures are selected considering the practical use of deepfake detection systems in social media. Given that real cases outnumber (deep)fake ones, we prioritized metrics related to False Positives (misclassifying real as fake) to prevent potential consequences such as suspicion, distrust, legal, or social repercussions, especially for users from specific ethnic groups.
Three fairness metrics are used to report the fairness performance of methods.
Specifically, we report the maximum differences in 
False Positive Rate (FPR) Gap ($G_{\text{FPR}}$) for Gender, Race, and Intersection groups. We also consider the Equal False Positive Rate ($F_{\text{FPR}}$) and Equal Odds ($F_{\text{EO}}$) metrics as used in \cite{wang2022understanding}. These metrics are defined as follows (for ease of notation, we write this for the training data but it is of course evaluated on test data):
\vspace{-1.5mm}
\begin{equation}\small
    \begin{aligned}
    G_{\text{FPR}} &\!:=\! \max_{g,g'\in\mathcal{G}} \big| \text{FPR}_g \!-\! \text{FPR}_{g'}\big|, \\
    F_{\text{FPR}} &\!:=\!\sum_{g\in \mathcal{G}}\Bigg|\frac{\sum_{i=1}^n\ind_{[\hat{Y}_i=1,G_i= g, Y_i=0]}}{\sum_{i=1}^n\ind_{[G_i= g, Y_i=0]}}-\frac{\sum_{i=1}^n\ind_{[\hat{Y}_i=1, Y_i=0]}}{\sum_{i=1}^n\ind_{[Y_i=0]}}\Bigg|, \\
    F_{\text{EO}} &\!:=\!\sum_{g\in \mathcal{G}}\sum_{q=0}^1\Bigg|\frac{\sum_{i=1}^n\ind_{[\hat{Y}_i=1,G_i= g, Y_i=q]}}{\sum_{i=1}^n\ind_{[G_i= g, Y_i=q]}}\!-\!\frac{\sum_{i=1}^n\ind_{[\hat{Y}_i=1, Y_i=q]}}{\sum_{i=1}^n\ind_{[Y_i=q]}}\Bigg|,
    \end{aligned}
\end{equation}
where $\text{FPR}_g$ represents the FPR scores of group $g$. %
$Y_i$ and $\hat{Y}_i$ respectively represent the true and predicted labels of the sample $X_i$. Their values are binary, where $0$ means real and $1$ means fake. For all fairness metrics here, lower is better.

Since there is usually a trade-off between fairness and detection performance~\cite{caton2020fairness,mehrabi2021survey}, we also include detection metrics to assess the balance between fairness and detection performance. Four widely-used deepfake detection metrics are reported: 1) the area under the curve (AUC), %
2) FPR, %
which is essential in real-world use as it indicates the count of incorrect fake classifications, 3) True Positive Rate (TPR), which measures the number of correct fake classification, and 4) Accuracy (ACC). We calculate the FPR, TPR, and ACC with a fixed threshold of 0.5~\cite{wang2023noise, cozzolino2023audio}. %

\smallskip
\noindent{\textbf{Baseline methods.}} We apply our proposed methods DAG-FDD and DAW-FDD in Section \ref{sec:method} to popular deepfake detectors to show their effectiveness. %
Five deepfake detection models are considered, including three widely-used CNN architectures in deepfake detection~\cite{li2020celeb, nadimpalli2022gbdf, nadimpalli2022improving} (\ie, Xception~\cite{rossler2019faceforensics++}, ResNet-50~\cite{he2016deep}, and EfficientNet-B3~\cite{tan2019efficientnet}) and two well-designed deepfake detectors with outstanding performance, namely DSP-FWA~\cite{li2020celeb} and RECCE~\cite{cao2022end}. We denote the detectors with their original loss functions (\eg, binary cross-entropy) as ``Original". %

\smallskip
\begin{table}[t]
\renewcommand\arraystretch{1}
\center 
\scalebox{0.48}{
\begin{tabular}{ c | c  c  c | c  c  c c | c | c}
 \hline
 \multirow{3}{*}{Methods} & \multicolumn{3}{c|}{Fairness Metrics (\%) \textdownarrow } & \multicolumn{4}{c|}{Detection Metrics (\%)} & \multirow{3}{*}{\makecell{{Training Time} \\ (mins)/Epoch}} & \multirow{3}{*}{\makecell{{Binary Search} \\ Time (mins)/Epoch}}\\ 
 \cline{2-8}
 & \multicolumn{3}{c|}{Intersection} & \multicolumn{4}{c|}{Overall} &  & \\ 
\cline{2-8}
 & $G_{\text{FPR}}$  &  $F_{\text{FPR}}$ & $F_{\text{EO}}$  & AUC \textuparrow & FPR \textdownarrow & TPR \textuparrow & ACC \textuparrow  &  & \\
\hline 
 \multirow{2}{*}{Original} & 24.00 & 45.50 & 63.24 & 95.00 & 19.28 & \textbf{95.94} & 93.29 & \multirow{2}{*}{2.6} & \multirow{2}{*}{N/A}\\ 
 & (9.00) & (16.39) & (12.96) & (2.96) & (7.14) & (1.41) & (1.60) &  & \\ 
 \hline
 \multirow{2}{*}{\makecell{{DAG-FDD} \\ {(Ours)}}} & \colorbox{gray!20}{13.83} & \colorbox{gray!20}{\textbf{24.38}} & \colorbox{gray!20}{48.52} & \colorbox{gray!20}{96.81} & \colorbox{gray!20}{13.43} & 95.33 & \colorbox{gray!20}{93.81} & \multirow{2}{*}{3.0}  & \multirow{2}{*}{0.59}\\
& (11.86) & (17.04) & (15.37) & (1.68) & (7.00) & (1.40) & (1.13) &  \\
\hline
 \multirow{2}{*}{\makecell{{DAW-FDD} \\ {(Ours)}}} &  \colorbox{gray!20}{\textbf{11.53}} & \colorbox{gray!20}{26.55} & \colorbox{gray!20}{\textbf{47.50}} & \colorbox{gray!20}{\textbf{97.40}} & \colorbox{gray!20}{\textbf{12.21}} & 95.45 & \colorbox{gray!20}{\textbf{94.11}} & \multirow{2}{*}{3.0} & \multirow{2}{*}{0.66}\\
  & (3.43) & (7.97) & (10.99) & (0.30) & (4.05) & (1.37) & (0.66) &  & \\
 \hline
\end{tabular}
}
\vspace{-0.3cm}
\caption{\label{tb:statistical} \textit{Detection mean and standard deviation (in parentheses) of Xception detector on FF++ testing set across 5 experimental repeats, in the same format as Table \ref{tb:baselines}. Training time and binary search time per epoch for each method are also reported.}}
\vspace{-6mm}
\end{table}

\begin{table*}[t]
\renewcommand\arraystretch{1}
\center
\scalebox{0.67}{
\begin{tabular}{ c | c | c  c  c | c  c  c |  c  c  c | c  c  c  c}
 \hline
 \multirow{3}{*}{Datasets} & \multirow{3}{*}{Methods} & \multicolumn{9}{c|}{Fairness Metrics (\%) \textdownarrow } & \multicolumn{4}{c}{Detection Metrics (\%)} \\ 
 \cline{3-15}
 & & \multicolumn{3}{c|}{Gender} & \multicolumn{3}{c|}{Race} & \multicolumn{3}{c|}{Intersection} & \multicolumn{4}{c}{Overall} \\ 
\cline{3-15}
& & $G_{\text{FPR}}$  & $F_{\text{FPR}}$  & $F_{\text{EO}}$  &  $G_{\text{FPR}}$  & $F_{\text{FPR}}$  & $F_{\text{EO}}$ & $G_{\text{FPR}}$  & $F_{\text{FPR}}$  & $F_{\text{EO}}$  & AUC \textuparrow & FPR \textdownarrow & TPR \textuparrow & ACC \textuparrow \\
\hline
\multirow{3}{*}{Celeb-DF} & Original & 4.93 & 4.93 & 22.04  & 3.31 & 4.77 & \textbf{26.06} & 11.81 & 15.66 & 39.95 & 97.17 & 13.01 & \textbf{95.83} & \textbf{94.05} \\ \cline{2-15} 
& DAG-FDD (Ours) & \colorbox{gray!20}{\textbf{2.02}} & \colorbox{gray!20}{\textbf{2.02}} & \colorbox{gray!20}{\textbf{16.77}}  & \colorbox{gray!20}{\textbf{1.20}} & \colorbox{gray!20}{\textbf{1.22}} &  {28.56} & \colorbox{gray!20}{\textbf{2.54}} &  \colorbox{gray!20}{\textbf{3.09}} & \colorbox{gray!20}{\textbf{30.43}} &  \colorbox{gray!20}{{98.00}} &  \colorbox{gray!20}{{2.42}} &  {87.40} &  {89.44} \\ 
& DAW-FDD (Ours) &  \colorbox{gray!20}{{3.81}} &  \colorbox{gray!20}{{3.81}} &  \colorbox{gray!20}{{18.93}}   &  \colorbox{gray!20}{{3.14}} &  \colorbox{gray!20}{{3.34}} & 33.91  &  \colorbox{gray!20}{{3.80}} &  \colorbox{gray!20}{{4.91}} &  \colorbox{gray!20}{{35.48}} & \colorbox{gray!20}{\textbf{98.03}} & \colorbox{gray!20}{\textbf{2.10}} & 84.53 & 87.21 \\ 
 \hline
 \multirow{3}{*}{DFD} & Original & 2.95 & 2.95 & 5.52 & 7.35 & 7.35 & 7.72 & 8.67 & 15.81 & 24.31 & 92.94 & \textbf{25.00} &  {96.01} & \textbf{89.09} \\ \cline{2-15}
& DAG-FDD (Ours) &  \colorbox{gray!20}{2.92} &  \colorbox{gray!20}{2.92} &  \colorbox{gray!20}{4.79} &   \colorbox{gray!20}{6.08} &  \colorbox{gray!20}{6.08} &  \colorbox{gray!20}{7.05} &  \colorbox{gray!20}{{8.30}} &  \colorbox{gray!20}{13.52} &  \colorbox{gray!20}{19.57} & \colorbox{gray!20}{\textbf{93.40}} & 28.07 & \colorbox{gray!20}{\textbf{96.31}} &  {88.28} \\ 
& DAW-FDD (Ours) & \colorbox{gray!20}{\textbf{1.40}} & \colorbox{gray!20}{\textbf{1.40}} & \colorbox{gray!20}{\textbf{3.14}}  & \colorbox{gray!20}{\textbf{2.36}} & \colorbox{gray!20}{\textbf{2.36}} & \colorbox{gray!20}{\textbf{3.35}} & \colorbox{gray!20}{\textbf{7.20}} & \colorbox{gray!20}{\textbf{8.74}} & \colorbox{gray!20}{\textbf{14.70}} &  \colorbox{gray!20}{{93.17}} &  {27.75} & 95.95 & 88.14 \\
 \hline
 \multirow{3}{*}{DFDC} & Original  & 1.64 & 1.64 & 4.36 & 4.02 & 5.85 & \textbf{38.84} & 20.17 & 38.68 & 119.71 & 92.40 & 7.28 & \textbf{76.32} & 86.87 \\ \cline{2-15}
& DAG-FDD (Ours) & \colorbox{gray!20}{\textbf{1.30}} & \colorbox{gray!20}{\textbf{1.30}} & 5.38  & 4.50 & \colorbox{gray!20}{5.78} & 46.56 & \colorbox{gray!20}{14.65} & \colorbox{gray!20}{33.79} & \colorbox{gray!20}{\textbf{113.93}} & \colorbox{gray!20}{92.69} & \colorbox{gray!20}{6.61} & 74.41 & 86.61  \\
& DAW-FDD (Ours) & 1.73 & 1.73  & \colorbox{gray!20}{\textbf{3.39}}  & \colorbox{gray!20}{\textbf{3.48}} & \colorbox{gray!20}{\textbf{4.14}} & 42.87 & \colorbox{gray!20}{\textbf{11.19}} & \colorbox{gray!20}{\textbf{23.63}} & \colorbox{gray!20}{115.15} & \colorbox{gray!20}{\textbf{94.88}} & \colorbox{gray!20}{\textbf{4.27}} & 75.10 & \colorbox{gray!20}{\textbf{88.37}} \\
 \hline
\end{tabular}
}
\vspace{-0.3cm}
\caption{\label{tb:datasets} \textit{
Results of Xception detector on Celeb-DF, DFD, and DFDC testing sets, in the same format as Table \ref{tb:baselines}.
}}
\vspace{-4mm}
\end{table*}

\begin{table*}[t]
\renewcommand\arraystretch{1}
\center
\scalebox{0.64}{
\begin{tabular}{ c | c |  c  c  c |  c  c  c |  c  c  c | c  c  c  c}
 \hline
 \multirow{3}{*}{Models} & \multirow{3}{*}{Methods} & \multicolumn{9}{c|}{Fairness Metrics (\%) \textdownarrow } & \multicolumn{4}{c}{Detection Metrics (\%)} \\ 
 \cline{3-15}
 & & \multicolumn{3}{c|}{Gender} & \multicolumn{3}{c|}{Race} & \multicolumn{3}{c|}{Intersection} & \multicolumn{4}{c}{Overall} \\ 
\cline{3-15}
& & $G_{\text{FPR}}$ & $F_{\text{FPR}}$ & $F_{\text{EO}}$  & $G_{\text{FPR}}$ & $F_{\text{FPR}}$ & $F_{\text{EO}}$ & $G_{\text{FPR}}$ & $F_{\text{FPR}}$ & $F_{\text{EO}}$ & AUC \textuparrow & FPR \textdownarrow & TPR \textuparrow & ACC \textuparrow \\
\hline
\multirow{3}{*}{ResNet-50} & Original  & 2.58 & 2.58 & 6.64 &  12.80 & 15.45 & 17.62  & 20.06 & 43.18 & 60.66 & 94.32 & 25.96 & \textbf{96.36} & \textbf{92.37} \\ \cline{2-15}
& \makecell{DAG-FDD (Ours)} & \colorbox{gray!20}{\textbf{2.21}} & \colorbox{gray!20}{\textbf{2.21}} & \colorbox{gray!20}{\textbf{6.37}} & \colorbox{gray!20}{\textbf{6.44}} & \colorbox{gray!20}{11.42} & \colorbox{gray!20}{14.17} & \colorbox{gray!20}{16.68} & \cellcolor{gray!20}{37.50} & \colorbox{gray!20}{\textbf{52.79}} & \colorbox{gray!20}{\textbf{94.51}} & \colorbox{gray!20}{\textbf{22.52}} & 95.34 & 92.15 \\
& DAW-FDD (Ours) & 3.78 & 3.78 & 10.21 &  \colorbox{gray!20}{7.01} & \colorbox{gray!20}{\textbf{8.27}} & \colorbox{gray!20}{\textbf{13.09}}  & \colorbox{gray!20}{\textbf{13.28}} & \cellcolor{gray!20}{\textbf{35.08}} & \colorbox{gray!20}{60.07} & 93.70 & \colorbox{gray!20}{23.56} & 93.65 & 90.58 \\
 \hline
\multirow{3}{*}{EfficientNet-B3} & Original & 1.97 & 1.97 & \textbf{4.15} &  9.05 & 10.86 & 14.12 & 13.38 & 22.65 & \textbf{40.13} & 95.91 & 20.25 & \textbf{97.21} & \textbf{94.09} \\
\cline{2-15}
& DAG-FDD (Ours)  & \colorbox{gray!20}{0.47} & \colorbox{gray!20}{0.47} & 5.36 & 9.48 & \colorbox{gray!20}{9.58} & \colorbox{gray!20}{13.50}  & \colorbox{gray!20}{10.87} & \cellcolor{gray!20}{19.34} & 46.08 & \colorbox{gray!20}{\textbf{97.20}} & \colorbox{gray!20}{8.40} & 92.87 & 92.65 \\
& DAW-FDD (Ours)  & \colorbox{gray!20}{\textbf{0.04}} & \colorbox{gray!20}{\textbf{0.04}} & 5.53 & \colorbox{gray!20}{\textbf{3.79}}  & \colorbox{gray!20}{\textbf{4.67}} & \colorbox{gray!20}{\textbf{12.63}}  & \colorbox{gray!20}{\textbf{6.43}} & \cellcolor{gray!20}{\textbf{12.57}} & 43.72 & \colorbox{gray!20}{96.30} & \colorbox{gray!20}{\textbf{8.22}} & 91.43 & 91.49 \\
 \hline
 \multirow{3}{*}{DSP-FWA} & Original & 5.90 & 5.90 & 11.81 & 11.07 & 14.58 & 21.98 & 21.38 & 48.20 & 75.91 & \textbf{91.79} & 31.64 & 93.17 & 88.74 \\ \cline{2-15}
& DAG-FDD (Ours) & \colorbox{gray!20}{4.64} & \colorbox{gray!20}{4.64} & \colorbox{gray!20}{\textbf{9.77}} & 12.52 & 18.04 & 25.03  & \colorbox{gray!20}{15.61} & \cellcolor{gray!20}{40.57} & \colorbox{gray!20}{\textbf{74.54}} & 91.47 & 32.35 & \colorbox{gray!20}{\textbf{93.70}} & \colorbox{gray!20}{\textbf{89.05}} \\ 
& DAW-FDD (Ours) &\colorbox{gray!20}{\textbf{3.02}} & \colorbox{gray!20}{\textbf{3.02}} & \colorbox{gray!20}{11.30}  & \colorbox{gray!20}{\textbf{5.75}} & \colorbox{gray!20}{\textbf{10.52}} & \colorbox{gray!20}{\textbf{19.34}}  & \colorbox{gray!20}{\textbf{12.84}} & \cellcolor{gray!20}{\textbf{36.05}} & \colorbox{gray!20}{75.73} & 90.84 & \colorbox{gray!20}{\textbf{30.43}} & 91.97 & 87.97 \\
 \hline
\multirow{3}{*}{RECCE} & Original & 0.87 & 0.87 & \textbf{3.14} & 18.81 & 27.65 & 30.07 & 30.26 & 67.38 & 80.34 & 98.05 & 21.20 & \textbf{98.21} & 94.74 \\ \cline{2-15}
& DAG-FDD (Ours)  & \colorbox{gray!20}{0.55} & \colorbox{gray!20}{0.55} & 3.71 & \colorbox{gray!20}{12.68} & \colorbox{gray!20}{17.41} & \colorbox{gray!20}{20.33} & \colorbox{gray!20}{15.40} & \colorbox{gray!20}{36.17} & \colorbox{gray!20}{54.24} & \colorbox{gray!20}{98.33} & \colorbox{gray!20}{12.01} & 96.80 & \colorbox{gray!20}{\textbf{95.23}} \\ 
& DAW-FDD (Ours)  & \colorbox{gray!20}{\textbf{0.25}} & \colorbox{gray!20}{\textbf{0.25}} & 4.75 & \colorbox{gray!20}{\textbf{6.99}} & \colorbox{gray!20}{\textbf{7.96}} & \colorbox{gray!20}{\textbf{11.95}} & \colorbox{gray!20}{\textbf{13.54}} & \colorbox{gray!20}{\textbf{23.44}} & \colorbox{gray!20}{\textbf{52.95}} & \colorbox{gray!20}{\textbf{98.35}} & \colorbox{gray!20}{\textbf{8.15}} & 94.59 & 94.10 \\
 \hline
\end{tabular}
}
\vspace{-0.3cm}
\caption{\label{tb:detectors} \textit{Results of ResNet-50, EfficientNet-B3, DSP-FWA, and RECCE detectors on FF++ testing set, in the same format as Table~\ref{tb:baselines}.}} 
\vspace{-6mm}
\end{table*}

In terms of comparison in fairness detection, we consider the method~\cite{nadimpalli2022gbdf} based on balancing the number of training samples in each group for deepfake fairness improvement; %
we take this method as a baseline for comparison denoted as ``Naive". Specifically, we use an intersectional group with the smallest number of training samples and then randomly select the same number of training samples from the other groups to create such a balanced training dataset. Moreover, we compare our two loss functions with the $\chi^2$-divergence based DRO ($\text{DRO}_{\chi^2}$)~\cite{hashimoto2018fairness}, the fairness risk measure (FRM) \cite{williamson2019fairness}, and a popular Group DRO method \cite{sagawa2019distributionally} in fairness research although they have not been applied to deepfake detection. {Besides, we modify $F_{\text{FPR}}$ and $F_{\text{EO}}$ as regularization terms~\cite{wang2022understanding}, and incorporate them with binary cross-entropy loss as baselines: $Cons$. EFPR and $Cons$. EO.
}

\smallskip
\vspace{-1mm}
\noindent{\textbf{Implementation details.}} All experiments are conducted on the PyTorch platform~\cite{paszke2019pytorch} using 4 NVIDIA RTX A6000 GPU cards. We train all methods by using a (mini-batch) stochastic gradient descent optimizer with batch size 640, epochs 200, and learning rate as $5\times10^{-4}$. %
We build our loss functions on the binary cross-entropy loss for the binary deepfake classification task. Since the DAW-FDD method needs to pre-define a set of groups, we use the Intersection group in experiments and also report the performance on single attributes. %
The hyperparameters $\alpha$ and $\alpha_g$ are tuned on the grid \{0.1, 0.3, 0.5, 0.7, 0.9\}. Following \cite{keya2021equitable}, the final hyperparameter setting per dataset and per method is determined based on a preset rule that allows up to a 5\% degradation of overall AUC in the validation set from the corresponding ``Original'' method while minimizing the intersectional $F_\text{FPR}$. More details and the evaluation of the influence of different parameter settings on detection performance are provided in Appendix \ref{appendix:sec:Hyperparameter-Tuning}, \ref{appendix:sensiparams}. 

\vspace{-1mm}
\subsection{Results}
\vspace{-2mm}
\noindent{\textbf{Performance on FF++ dataset.}} We first report results of our methods compared with several baselines on the FF++ dataset using the Xception deepfake detector in Table~\ref{tb:baselines}. These results show that in the cases where demographic information is unavailable from the training data, our DAG-FDD method achieves superior fairness performance to the Original method across most metrics for all three sensitive attribute groups, as shown in gray highlights. For example, we enhance the $G_{\text{FPR}}$ of Gender, $F_{\text{EO}}$ of Race, and $F_{\text{FPR}}$ of Intersection by 2.47\%, 9.51\%, and 10.38\%, respectively, compared to the Original. 
These results indicate our method has strong applicability in scenarios where demographic data is unavailable. In addition, it is clear that our method outperforms the $\text{DRO}_{\chi^2}$ method on most fairness metrics. 
This is benefited from the tighter upper bound (see Proposition \ref{prop:cvar-risk}) on the risk $\mathcal{R}_{\max}(\theta)$ in our DAG-FDD method than the $\text{DRO}_{\chi^2}$ method as mentioned in \cite{zhai2021doro}.

With demographic information, we see that the Naive method trained on a balanced dataset does not guarantee an improvement in fairness metrics on test data. For example, on the intersectional groups, all fairness scores of the Naive method (\eg, $F_{\text{FPR}}$: 72.19\%)  are worse than the Original method (\eg, $F_{\text{FPR}}$: 31.59\%).  
This can be attributed to the fact that a naive balancing strategy will reduce the number of available training samples, resulting in a significant decrease in detection performance. The same trends can be found in AUC scores, which decrease from 92.76\% (Original) to 83.17\% (Naive), and in FPR scores, which increase from 22.06\% (Original) to 50.77\% (Naive). Thus, a poorly trained model on balanced data could result in worse fairness scores. %

\begin{figure*}[!t] 
\centering 
\includegraphics[width=0.92\textwidth]{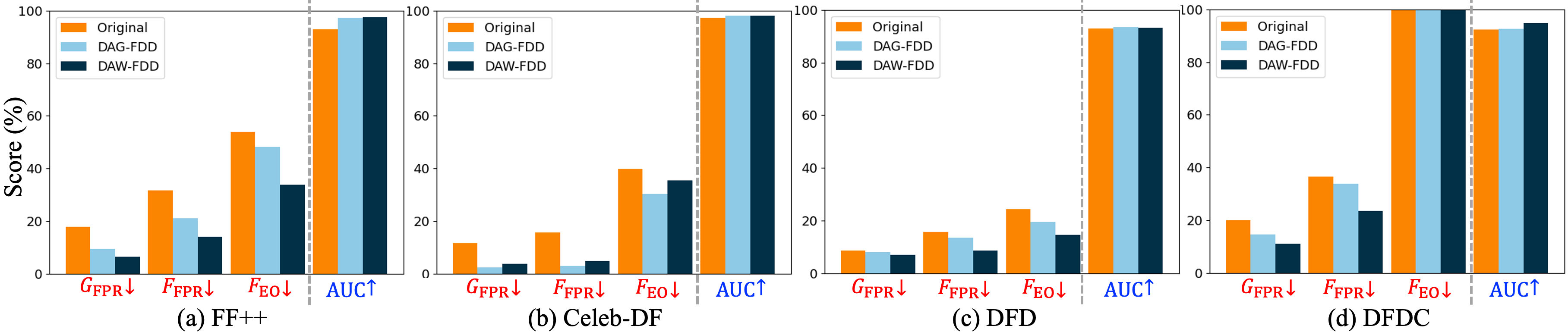}
\vspace{-3.5mm}
\caption{\textit{Comparison of Xception detector on Intersection group of four datasets: (a) FF++, (b) DFDC, (c) DFD, and (d) DFDC.}}%
\label{fig:baron4datasets} 
\vspace{-3.5mm}
\end{figure*}

\begin{figure*}[!t] 
\centering 
\includegraphics[width=0.92\textwidth]{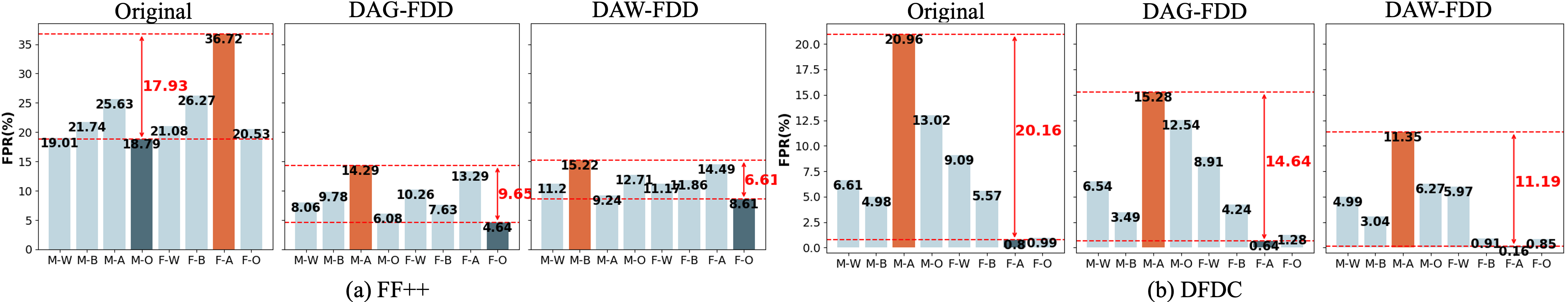}
\vspace{-3.5mm}
\caption{\textit{FPR comparison of Xception detector on intersectional groups of (a) FF++ and (b) DFDC dataset. \colorbox{yancolor1!80}{Orange} and \colorbox{yancolor2!80}{dark cyan} bars show the groups with the highest and lowest FPR, respectively, while the double arrow indicates their gap (smaller is better).}} 
\label{fig:subgroupbar} 
\vspace{-6mm}
\end{figure*}

Our DAW-FDD method outperforms all methods on the most fairness metrics (as shown in Bold). The reason is that DAW-FDD uses the additional demographic information to guide training to achieve fairness without reducing the dataset size. %
In particular, the DAW-FDD method achieves the best fairness performance on all metrics in the Intersection group thanks to the guidance of intersection group information in the design of DAW-FDD. %
The superiority of DAW-FDD over Group DRO is evident. 
Specifically, Group DRO does not show any improvement, possibly because it places greater emphasis on improving the worst-group generalization performance and less on ensuring overall fairness. 
Furthermore, in our comparison between the DAW-FDD and FRM methods, we have found our method outperforms FRM. This result clearly demonstrates the effectiveness of our learning strategy that considers two types of imbalance (demographic groups and, separately, real vs deepfake).

Most importantly, our DAG-FDD and DAW-FDD methods not only enhance fairness performance but also improve the detection performance of the detector. 
For example, we see improvements of approximately 4.7\% in AUC and 12.52\% in FPR when compared with the Original method. To show the stability of our methods, we run experimental repeats with 5 random seeds as shown in Table~\ref{tb:statistical}. It is clear that our methods robustly improve fairness. {%
We also show the training time per epoch costs during training Xception on FF++ dataset in Table~\ref{tb:statistical}. Based on the presented table results, our methods show a slightly higher time requirement compared to the original method. However, the difference is minimal, mainly due to the incorporation of a binary search in the calculation of model training loss.}

\noindent{\textbf{Performance on different datasets.}} Table~\ref{tb:datasets} shows the evaluation performance of the Xception detector on three popular deepfake datasets. It is clear that our proposed DAG-FDD and DAW-FDD methods outperform the Original method on all three datasets across all groups and most fairness metrics, especially on the Intersection group (also as shown in Figure \ref{fig:baron4datasets}). Moreover, our methods achieve similar or better scores on most detection metrics. %
Note that our methods on the Celeb-DF dataset lead to a decrease in TPR. One possible reason is that our methods involve hyperparameter tuning based on $F_{\text{FPR}}$, as mentioned in experimental settings. To evaluate the effectiveness of our method on other metrics, we employ $F_\text{EO}$ as an index to tune the hyperparameter and report the results in Appendix~\ref{appendix:set:opbyeo}. The results illustrate that optimizing hyperparameters using $F_\text{EO}$ can improve TPR and $F_\text{EO}$.
This demonstrates the good flexibility and applicability of our methods to different metrics and datasets.
We further show the FPR comparison results on FF++ and DFDC datasets with detailed performance in groups in Figure \ref{fig:subgroupbar}.
Our methods evidently narrow the disparity between groups and lower the FPR of each group.

\noindent{\textbf{Performance on various detection models.}} We further evaluate the effectiveness of our methods on four popular deepfake detection models on the FF++ dataset. The results are presented in Table~\ref{tb:detectors}. It is clear that our methods can improve the fairness performance of the detectors without significantly decreasing the detection performance. These results indicate that our methods exhibit high scalability and can be seamlessly integrated with different backbones and deepfake detection models.

\vspace{-2mm}
\section{Conclusion}\vspace{-1mm}
In this work, we propose two methods, DAG-FDD and DAW-FDD, for training fair deepfake detection models in ways that are agnostic to or, separately, aware of demographic factors. Extensive experiments on four large-scale deepfake datasets and five deepfake detectors show the effectiveness of our methods in improving the fairness of existing deepfake detectors.

A limitation of our methods is that they rely on the assumption that loss functions can be decomposed into individual terms and that each instance is independent. Therefore, integrating our methods into graph learning-based detectors may not be straightforward.

In future work, we aim to extend this work in the following areas. First, we will examine the fairness generalization abilities of cross-dataset deepfake detection. Second, we will investigate fairness methods for managing non-decomposable loss-based detectors.

\noindent \textbf{Acknowledgement.} This work was supported in part by the US
Defense Advanced Research Projects Agency (DARPA) Semantic Forensic (SemaFor) program, under Contract No.
HR001120C0123. G. H. Chen is supported by NSF \mbox{CAREER} award \#2047981.

\newpage
{\small
\bibliographystyle{ieeetr}
\bibliography{egbib}
}

\newpage
\input{Appendix}

\end{document}

%% file: Appendix.tex
\onecolumn
\appendix
\numberwithin{equation}{section}
\numberwithin{theorem}{section}
\numberwithin{figure}{section}
\numberwithin{table}{section}
\renewcommand{\thesection}{{\Alph{section}}}
\renewcommand{\thesubsection}{\Alph{section}.\arabic{subsection}}
\renewcommand{\thesubsubsection}{\Roman{section}.\arabic{subsection}.\arabic{subsubsection}}

\def\p{\mathbf{p}}
\def\v{\mathbf{v}}
\def\u{\mathbf{u}}

\begin{center}
\textbf{\Large Appendix for ``Improving Fairness in Deepfake Detection"}
\end{center}

\bigskip 

This Appendix provides proof of the proposed methods, additional experimental details and results. Specifically, Sections~\ref{sec1:proof}, \ref{appendix:algorithms}, and \ref{appendix:explicit_forms} provide proof and details of the proposed methods. Section~\ref{appendix:sec:Additional-Experimental-Details} provides details of our experiment, including parameter setting and source code. Section~\ref{appendix:sec:Additional-Experimental-Results} provides further analysis of additional experimental results, including optimization by different metric (in \ref{appendix:set:opbyeo}), effect of choosing different hyperparameters (in \ref{appendix:sensiparams}), performance on Cross-domain Dataset (in \ref{appendix:cross-domaintest}), convergence analysis of the proposed methods (in \ref{appendix:lossconvergence}), more comparison results~(in~\ref{appendix:reccecomparison} and \ref{appendix:dfplatter}), and details on datasets and results of each subgroup (in \ref{appendix:numberoftraining2} and \ref{appendix:details}). 

\section{Proofs}\label{sec1:proof}

\subsection{Proof of Proposition \ref{prop:cvar-risk}}\label{appendix:prop cvar-risk}

\begin{proof}
For any $m$, denote $Z=(X,Y)$ and $\mathcal{D}_m$ as a set that contains samples from $m$-th group, then $\mathbb{P}(Z)=\pi_m \mathbb{P}(Z|\mathcal{D}_m)+ (1-\pi_m)\mathbb{P}(Z|\overline{\mathcal{D}_m})$, where $\overline{\mathcal{D}_m}$ contains samples are not in $\mathcal{D}_m$. Let $\mathbb{Q}(Z)= \mathbb{P}(Z|\mathcal{D}_m)$ and $\mathbb{Q}'(Z)=\frac{\pi_m-\alpha}{1-\alpha} \mathbb{P}(Z|\mathcal{D}_m)+\frac{1-\pi_m}{1-\alpha}\mathbb{P}(Z|\overline{\mathcal{D}_m})$. Then $\mathbb{P}(Z) = \alpha \mathbb{Q}(Z) + (1-\alpha) \mathbb{Q}'(Z)$, which implies that 

\begin{equation*}
    \begin{aligned}
    \alpha\mathbb{E}_{\mathbb{Q}(Z)}[\ell(\theta;Z)-\lambda]=\mathbb{E}_{\alpha \mathbb{Q}(Z)}[\ell(\theta;Z)-\lambda]\leq\mathbb{E}_{\alpha \mathbb{Q}(Z)}[[\ell(\theta;Z)-\lambda]_+]\leq \mathbb{E}_{\mathbb{P}(Z)}[[\ell(\theta;Z)-\lambda]_+].
    \end{aligned}
\end{equation*}
The last inequality holds because $\alpha\leq \min_{m=1,...,K}\pi_m$ and $\alpha\in(0,1)$, which means $\mathbb{Q}'(Z)\geq 0$ and therefore $\mathbb{P}(Z)\geq \alpha \mathbb{Q}(Z)$.
From the above inequations, we obtain 
\begin{equation*}
    \begin{aligned}
    \alpha\mathbb{E}_{\mathbb{Q}(Z)}[\ell(\theta;Z)-\lambda] &\leq \mathbb{E}_{\mathbb{P}(Z)}[[\ell(\theta;Z)-\lambda]_+]\\
    \Rightarrow  \mathbb{E}_{\mathbb{Q}(Z)}[\ell(\theta;Z)-\lambda] &\leq \frac{1}{\alpha}\mathbb{E}_{\mathbb{P}(Z)}[[\ell(\theta;Z)-\lambda]_+]\\
    \Rightarrow  \mathbb{E}_{\mathbb{Q}(Z)}[\ell(\theta;Z)] & \leq \lambda+\frac{1}{\alpha}\mathbb{E}_{\mathbb{P}(Z)}[[\ell(\theta;Z)-\lambda]_+]= \text{CVaR}_\alpha(\theta)
    \end{aligned}
\end{equation*}

In Section \ref{sec:FDD-without-demographics}, we have already defined $\mathbb{P}_m$, which is just $\mathbb{Q}(Z)$.  Therefore, we have $\mathcal{R}_{\max}(\theta)\leq \text{CVaR}_\alpha(\theta)$.  

\end{proof}

\subsection{Proof of Theorem \ref{theorem:cvar-atk}}\label{proof:theorem:cvar-atk}
\begin{proof}

1) We first prove that $\frac{1}{k}\sum_{i=1}^k \ell_{[i]}=\min_{\lambda\in \mathbb{R}} \lambda + \frac{1}{k}\sum_{i=1}^q [\ell_i-\lambda]_+$.

\noindent
$\Rightarrow$: Suppose $\overline{\ell}:=\{\ell_1,...\ell_q\}$. We know $\sum_{i=1}^k \ell_{[i]}$ is the solution of 
\begin{equation*}
    \max_\p \p^\top \overline{\ell}, ~\text{s.t.}~ \p^\top \mathbf{1}=k, \mathbf{0}\leq\p\leq \mathbf{1}.
\end{equation*}
We apply Lagrangian to this equation and get
\begin{equation*}
    \mathcal{L} = -\p^\top \overline{\ell}-\v^\top\p+\u^\top(\p-1)+\lambda(\p^\top\mathbf{1}-k)
\end{equation*}
where $\u\geq\mathbf{0}$, $\v\geq\mathbf{0}$ and $\lambda\in\mathbb{R}$ are Lagrangian multipliers. Taking its derivative with respect to $\p$ and set it to 0, we have $\v=\u-\overline{\ell}+\lambda\mathbf{1}$. Substituting it back into the Lagrangian, we get 
\begin{equation*}
    \min_{\u,\lambda} \u^\top \mathbf{1}+k\lambda, ~\text{s.t.}~ \u\geq\mathbf{0}, \u+\lambda\mathbf{1}-\overline{\ell}\geq 0.
\end{equation*}
This means 
\begin{equation*}
    \sum_{i=1}^k \ell_{[i]}=\min_{\lambda}k\lambda+\sum_{i=1}^q[\ell_i-\lambda]_+.
\end{equation*}
Therefore, 
\begin{equation}
    \frac{1}{k}\sum_{i=1}^k \ell_{[i]}=\min_{\lambda}\lambda+\frac{1}{k}\sum_{i=1}^q[\ell_i-\lambda]_+.
\label{eq:topk_convex}
\end{equation}

\noindent
$\Leftarrow$: Denote $\overline{\mathcal{L}}:=\lambda+\frac{1}{k}\sum_{i=1}^q[\ell_i-\lambda]_+$. Since $\overline{\mathcal{L}}$ is a convex function with respect to $\lambda$, we can set the $\partial_{\lambda} \overline{\mathcal{L}}=0$ to get the optimal value of $\lambda^*$. Thus, we have $\partial_{\lambda} \overline{\mathcal{L}} = 1-\frac{1}{k} \sum_{i=1}^q \ind_{[\ell_i\geq\lambda^*]}=0$, then $\lambda^*= \ell_{[k]}$ can be an optimal value. Taking  $\lambda^*= \ell_{[k]}$ into $\overline{\mathcal{L}}$, we obtain $\overline{\mathcal{L}}=\frac{1}{k}\sum_{i=1}^k\ell_{[i]}$.

Based on the above analysis, we get $\frac{1}{k}\sum_{i=1}^k \ell_{[i]}=\min_{\lambda\in \mathbb{R}} \{\lambda + \frac{1}{k}\sum_{i=1}^q [\ell_i-\lambda]_+\}$.

2) Using the above result, we can directly replace $\mathcal{L}_g(\theta) = \frac{1}{k_g}\sum_{j=1}^{k_g}\ell_{[j]}^{g}(\theta)$ from (\ref{eq:DAW-FDD-ori}) with $\mathcal{L}_g(\theta)= \min_{\lambda_g\in \mathbb{R}}\{\lambda_g+\frac{1}{\alpha_gn_g}\sum_{i\in \mathcal{I}_g}[\ell(\theta;X_i,Y_i)-\lambda_g]_+\}$. This is also shown in (\ref{eq:eq:DAW-FDD}).

\end{proof}

\section{Pseudocode of the DAW-FDD}\label{appendix:algorithms}

\begin{algorithm}[h]\footnotesize
    \caption{DAW-FDD}\label{alg:DAW-FDD}
    \SetAlgoLined
    \KwIn{A training dataset $\mathcal{S}$ with demographic variable $G$, A set of subgroups $\mathcal{G}$,  $\alpha$, $\alpha_g$, max\_iterations, num\_batch, $\eta$}
    \KwOut{A fair deepfake detection model with parameters $\theta^*$} 
    
    \textbf{Initialization:} $\theta_0$, $l=0$

    \For{$e=1$ to \emph{max\_iterations}}{
    \For{$b=1$ to \emph{num\_batch}}{
    { \mbox{Sample a mini-batch $\mathcal{S}_b$ from $\mathcal{S}$}}
    
    Compute $\ell(\theta_l;X_i,Y_i)$, $\forall (X_i,Y_i)\in \mathcal{S}_b$
    
    For each $g\in\{1,...,|\mathcal{G}|\}$, set $\lambda_g^*$ to be the value of $\lambda_g$ that minimizes $\mathcal{L}_g(\theta,\lambda_g)$ as given in (\ref{eq:eq:DAW-FDD2}). This minimization is solved using binary search.
    
    Set $L_g(\theta)\leftarrow L_g(\theta, \lambda_g^*)$ using (\ref{eq:eq:DAW-FDD2}), $\forall g$
    
    Using binary search to find $\lambda$ that minimizes (\ref{eq:eq:DAW-FDD1})
    
    Set $\theta_{l+1}\leftarrow \theta_l-\eta \partial_{\theta}\mathcal{L}_{\text{DAW-FDD}}(\theta_l,\lambda)$ 
    
    $l\leftarrow l+1$
    
    }
    }
    \Return{$\theta^* \leftarrow \theta_{l}$}
    \vspace{-1mm}
\end{algorithm}

\section{Explicit Forms of (sub) gradients}\label{appendix:explicit_forms}
From equation (\ref{eq:eq:DAW-FDD}), we have 
\begin{equation*}
    \begin{aligned}
    &\min_{\theta\in\Theta,\lambda\in \mathbb{R}} \mathcal{L}_{\emph{DAW-FDD}}(\theta,\lambda)\!:=\! \lambda+\frac{1}{\alpha |\mathcal{G}|}\sum_{g\in\mathcal{G}}
      [\mathcal{L}_g(\theta)-\lambda]_+, \\ 
&\emph{s.t.} \ \mathcal{L}_g(\theta)\!\!=\!\! \min_{\lambda_g\in \mathbb{R}}\!\mathcal{L}_g(\theta,\lambda_g)\!:=\!\lambda_g\!+\!\frac{1}{\alpha_g n_g}\!\!\sum_{i\in\mathcal{I}_g}\! [\ell(\theta;X_i,Y_i)-\lambda_g]_+.
    \end{aligned}
\end{equation*}
We can get
\begin{equation*}
    \begin{aligned}
    \partial_{\theta}\mathcal{L}_{\text{DAW-FDD}}(\theta,\lambda)= \frac{1}{\alpha |\mathcal{G}|} \sum_{g\in\mathcal{G}}\Bigg[\Bigg(\frac{1}{\alpha_gn_g} \sum_{i\in\mathcal{I}_g}\partial_{\theta}\ell(\theta;X_i,Y_i)\cdot \ind_{[\ell(\theta;X_i,Y_i)>\lambda_i]}\Bigg)\cdot \ind_{[\mathcal{L}_g(\theta)>\lambda]}\Bigg]
    \end{aligned}
\end{equation*}

\section{Additional Experimental Details}\label{appendix:sec:Additional-Experimental-Details}

\subsection{$\alpha$ and $\alpha_g$ Settings on Each Dataset}\label{appendix:sec:Hyperparameter-Tuning}
We tune $\alpha$ and $\alpha_g$ on the following hyperparameter grid: 0.1, 0.3, 0.5, 0.7, 0.9. We provide a reference for setting $\alpha$ and $\alpha_g$ to reproduce our experimental results in Table~\ref{tb:alphasetting}.

\begin{table}[!h]
\renewcommand\arraystretch{1}
\center 
\scalebox{0.83}{
\begin{tabular}{ c | c | c |  c | c | c | c | c | c }
 \hline
 \multirow{2}{*}{Parameter} & \multicolumn{4}{c|}{Xception} & {ResNet-50} & {EfficientNet-B3} & {DSP-FWA} & {RECCE} \\ 
 \cline{2-9}
 & FF++ & Celeb-DF & DFD & DFDC & FF++ & FF++ & FF++ & FF++ \\
\hline 
$\alpha$ in DAG-FDD  & 0.5 & 0.5 & 0.3 & 0.7 & 0.5 & 0.5 & 0.5 & 0.5 \\
 \hline
 $\alpha,\alpha_{g}$ in DAW-FDD & 0.5, 0.9 & 0.5,0.7 & 0.7, 0.9& 0.5, 0.7 & 0.5, 0.9 & 0.5, 0.9 & 0.7, 0.9 & 0.5, 0.9\\
\hline
\end{tabular}
}
\caption{\label{tb:alphasetting} \textit{
Hyperparameter settings of DAG-FDD and DAW-FDD.}}
\end{table}

\subsection{Trade-off Parameters for $Cons$. EFPR and $Cons$. EO}

For $Cons$. EFPR and $Cons$. EO baselines, we tune the trade-off hyperparameters on the following grid: 0.5, 0.6, 0.7, 0.8, 0.9. Finally, we use 0.6 for both methods since this hyperparameter can return the best performance.

\section{Additional Experimental Results}\label{appendix:sec:Additional-Experimental-Results}

\subsection{Optimization by Metric $F_\text{EO}$}\label{appendix:set:opbyeo}
We employ $F_\text{EO}$ as an index to tune the hyperparameter and report the results in Tabel~\ref{tb:metriceo}. The results illustrate that optimizing hyperparameters using $F_\text{EO}$ can improve TPR and $F_\text{EO}$ (compared with results in Table \ref{tb:datasets}), which demonstrates that our method can generalize to different metric.

\begin{table}[!h]
\renewcommand\arraystretch{1}
\center
\scalebox{0.75}{
\begin{tabular}{ c | c  c  c  c | c  c  c  c}
 \hline
 \multirow{3}{*}{Methods} & \multicolumn{4}{c|}{Fairness Metrics (\%) \textdownarrow } & \multicolumn{4}{c}{Detection Metrics (\%)} \\ 
 \cline{2-9}
 & \multicolumn{4}{c|}{Intersection} & \multicolumn{4}{c}{Overall} \\ 
\cline{2-9}
& $G_{\text{AUC}}$  & $G_{\text{FPR}}$  & $F_{\text{FPR}}$  & $F_{\text{EO}}$  & AUC \textuparrow & FPR \textdownarrow & TPR \textuparrow & ACC \textuparrow \\
\hline
Original & 8.53 & 11.81 & 15.66 & 39.95 & 97.17 & 13.01 & \textbf{95.83} & \textbf{94.05} \\
\hline
DAG-FDD (Ours) & \colorbox{gray!20}{\textbf{5.86}} & \colorbox{gray!20}{7.04} & \colorbox{gray!20}{8.23} & \colorbox{gray!20}{\textbf{29.65}} & \colorbox{gray!20}{98.50} & \colorbox{gray!20}{5.06} & 93.48 & 93.78 \\ 
DAW-FDD (Ours) & \colorbox{gray!20}{6.67} & \colorbox{gray!20}{\textbf{2.96}} & \colorbox{gray!20}{\textbf{3.96}} & \colorbox{gray!20}{30.52} & \colorbox{gray!20}{\textbf{98.81}} & \colorbox{gray!20}{\textbf{2.78}} & 91.99 & 93.04 \\
 \hline
\end{tabular}
}
\caption{\label{tb:metriceo} \textit{
Test set results of Xception on the Celeb-DF dataset, optimized by $F_\text{EO}$ metric.
}} 
\end{table}

\subsection{Effect of $\alpha$ and $\alpha_g$}\label{appendix:sensiparams}

\begin{figure}[!h] 
\centering 
\includegraphics[width=0.75\textwidth]{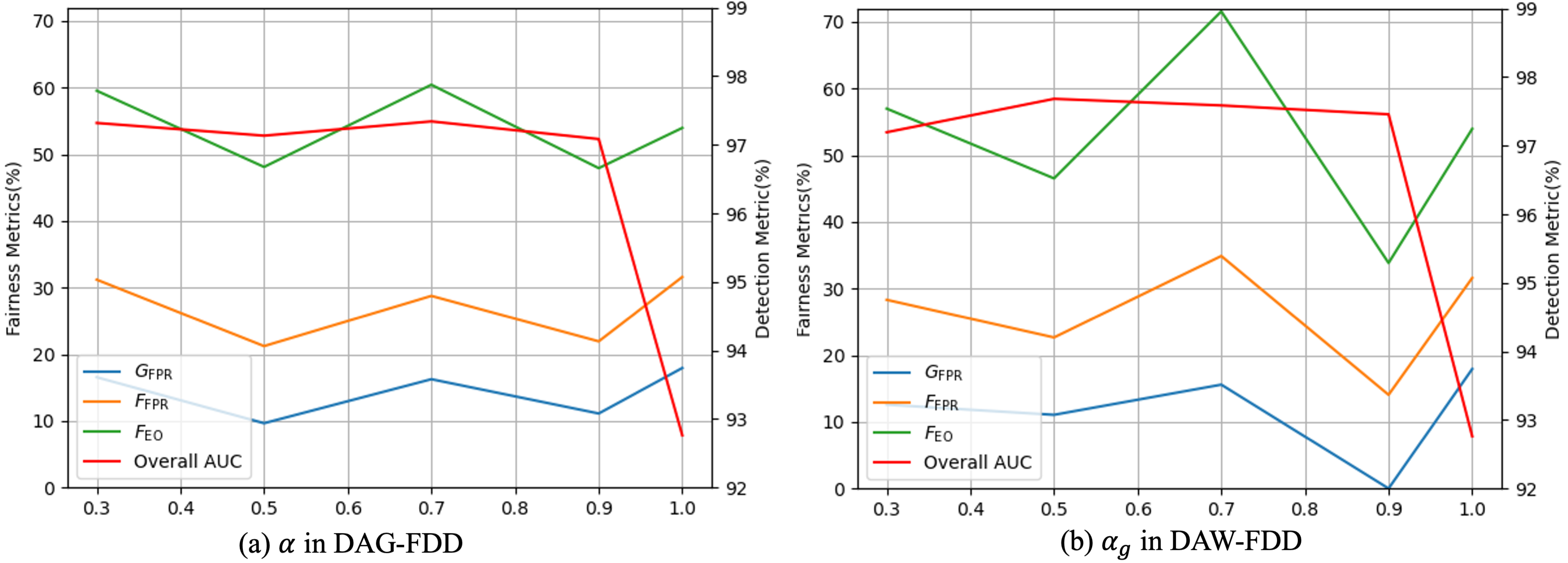}
\caption{\textit{Parameters of DAG-FDD and DAW-FDD on FF++ dataset with Xception.}} 
\label{fig:ablation} 
\end{figure}

Fig.~\ref{fig:ablation} shows the fairness metrics and performance metric AUC to different $\alpha$ and $\alpha_{g}$ values in DAG-FDD and DAW-FDD methods, respectively, when using Xception as backbone in FF++ dataset. Experiment result in Fig.~\ref{fig:ablation} (a) demonstrates that the model achieves the best fairness performance when setting $\alpha$ as $0.5$ in DAG-FDD and also keeps fair AUC score. In FAW-FDD, we set $\alpha$ as 0.5 selected from the range of $\{0.1, 0.3, 0.5, 0.7, 0.9\}$ based on the best fairness performance first. Secondly, we searched for the optimal value of $\alpha_g$ in the range of $\{0.1, 0.3, 0.5, 0.7, 0.9\}$ while keeping $\alpha$ fixed at its optimal value. Fig.~\ref{fig:ablation} (b) shows that the proposed DAW-FDD performs best when $\alpha_g$ is set to $0.9$ when $\alpha$ is fixed on 0.5.

\subsection{Performance on Cross-domain Dataset}\label{appendix:cross-domaintest}

We further evaluate the performance of our methods using Xception on cross-domain dataset. The models are trained on FF++ dataset and tested on DFDC dataset. The results are presented in Table~\ref{tb:crossdataset}. 

\begin{table}[!h]
\renewcommand\arraystretch{1}
\center 
\scalebox{0.8}{
\begin{tabular}{ c | c  c  c  c | c  c  c  c}
 \hline
 \multirow{3}{*}{Methods} & \multicolumn{4}{c|}{Fairness Metrics (\%) \textdownarrow } & \multicolumn{4}{c}{Detection Metrics (\%)} \\ 
 \cline{2-9}
 & \multicolumn{4}{c|}{Intersection} & \multicolumn{4}{c}{Overall} \\ 
\cline{2-9}
 & $G_{\text{AUC}}$  & $G_{\text{FPR}}$  &  $F_{\text{FPR}}$  & $F_{\text{EO}}$  & AUC \textuparrow & FPR \textdownarrow & TPR \textuparrow & ACC \textuparrow \\
\hline 
 Original & 33.76 & 17.19 & 30.70 & 122.51 & 58.81 & 59.54 & 71.60 & 51.57 \\ 
 \hline
 DAG-FDD (Ours) & 25.42 & 24.16 & 49.27 & 117.19 & 56.32 & 35.29 & 47.06 & 58.41 \\
\hline
 DAW-FDD (Ours) & 26.96 & 21.50 & 45.34 & 119.32 & 59.95 & 43.70 & 60.69 & 57.87 \\
 \hline
\end{tabular}
}
\caption{\label{tb:crossdataset} \textit{Cross-domain Performance. Models are trained on FF++ and tested on DFDC.}}
\vspace{-5mm}
\end{table}

\subsection{Convergence of the Proposed Loss Functions} \label{appendix:lossconvergence}
\begin{figure}[!h] 
\centering 
\includegraphics[width=0.4\textwidth]{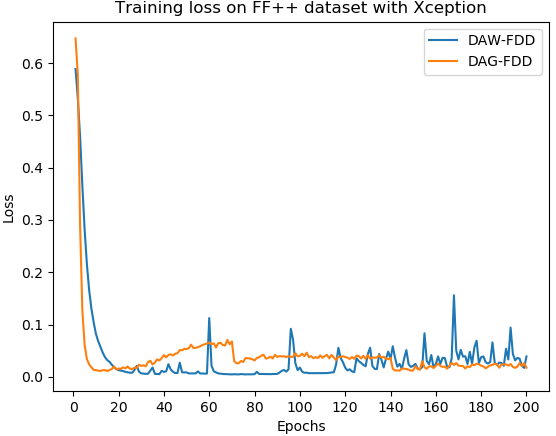}
\caption{\textit{Training loss convergence.}} 
\label{fig:trainloss} 
\end{figure}

We also show the training loss convergence of our methods when applying to Xception on FF++ dataset in Fig.~\ref{fig:trainloss}. The results show that our methods can converge within reasonable epochs. 

\subsection{Comparison on SOTA Deepfake Detector}
\label{appendix:reccecomparison}
Since the RECCE model achieves SOTA detection performance on several datasets, we apply our methods and baselines based on the RECCE detector and show the results in Table~\ref{tb:detailedrecce}. The results demonstrate the adaptability and efficiency of our methods.

\begin{table*}[t]
\renewcommand\arraystretch{1}
\center
\scalebox{0.75}{
\begin{tabular}{c | c |  c  c  c |  c  c  c |  c  c  c | c  c  c  c}
 \hline
 \multirow{3}{*}{Methods} & \multirow{3}{*}{\makecell{Require\\Demo-\\graphics}} & \multicolumn{9}{c|}{Fairness Metrics (\%) \textdownarrow } & \multicolumn{4}{c}{Detection Metrics (\%)} \\ 
 \cline{3-15}
& & \multicolumn{3}{c|}{Gender} & \multicolumn{3}{c|}{Race} & \multicolumn{3}{c|}{Intersection} & \multicolumn{4}{c}{Overall} \\ 
\cline{3-15}
& & $G_{\text{FPR}}$  & $F_{\text{FPR}}$  & $F_{\text{EO}}$  & $G_{\text{FPR}}$  & $F_{\text{FPR}}$  & $F_{\text{EO}}$ & $G_{\text{FPR}}$  &  $F_{\text{FPR}}$  & $F_{\text{EO}}$  & AUC \textuparrow & FPR \textdownarrow & TPR \textuparrow & ACC \textuparrow \\
\hline 
Original & $-$ &  0.87 & 0.87 & \textbf{3.14} & 18.81 & 27.65 & 30.07 & 30.26 & 67.38 & 80.34 & 98.05 & 21.20 & \textbf{98.21} & 94.74 \\ \hline \hline
$\text{DRO}_{\chi^2}$~\cite{hashimoto2018fairness} & \multirow{2}{*}{$\times$} & 1.46 & 1.46 & 4.17  & 13.87 &  20.05 & 23.15 & 23.63 & 42.14 & 57.29 & 98.32 & 15.65 & 97.28 &  94.97  \\  

 \makecell{DAG-FDD (Ours)} &   & \colorbox{gray!20}{0.55} & \colorbox{gray!20}{0.55} & 3.71  & \colorbox{gray!20}{12.68} & \colorbox{gray!20}{17.41} & \colorbox{gray!20}{20.33}  & \colorbox{gray!20}{15.40} & \colorbox{gray!20}{36.17} & \colorbox{gray!20}{54.24} & \colorbox{gray!20}{98.33} & \colorbox{gray!20}{12.01} & 96.80 & \colorbox{gray!20}{\textbf{95.23}}  \\
\hline \hline
Naive~\cite{nadimpalli2022gbdf}  & \multirow{6}{*}{$\checkmark$}  & 9.48 & 9.48 & 13.05  & 18.26 & 20.86 & 22.27 & 28.74 & 73.59 & 89.87 & 93.64 & 27.57 & 95.96 &  91.76 \\
FRM~\cite{williamson2019fairness} &  & 2.15 & 2.15 & 5.48  & 8.50 & 10.00 & 13.75 & 14.88 & 30.59 & \textbf{49.86} & 98.06 & 15.21 & 97.05 &  94.86  \\ 
Group DRO~\cite{sagawa2019distributionally} & & 0.74 & 0.74 & 3.71  & 12.08 & 16.26 & 20.01 & 15.17 & 32.95 & 51.08 & 98.22 & 11.75 & 96.59 &  95.10 \\
$Cons$. EFPR~\cite{wang2022understanding} & &  6.13 & 6.13 & 11.15  & 10.71 & 15.00 & 19.46 & 13.67 & 38.48 & 63.80 & 97.17 & 14.72 & 96.29 &  94.32  \\
DAW-FDD (Ours) &  & \colorbox{gray!20}{\textbf{0.25}} & \colorbox{gray!20}{\textbf{0.25}} & 4.75  & \colorbox{gray!20}{\textbf{6.99}} & \colorbox{gray!20}{\textbf{7.96}} & \colorbox{gray!20}{\textbf{11.95}} &  \colorbox{gray!20}{\textbf{13.54}} & \colorbox{gray!20}{\textbf{23.44}} & 52.95 & \colorbox{gray!20}{\textbf{98.35}} & \colorbox{gray!20}{\textbf{8.15}} & 94.59 & 94.10  \\

 \hline
\end{tabular}
}
\caption{\label{tb:detailedrecce} \textit{Comparison results with different fairness solutions using RECCE Deepfake detector on FF++ testing set across Gender, Race, and Intersection groups. The best results are shown in \textbf{Bold}. $\uparrow$ means higher is better and $\downarrow$ means lower is better. \colorbox{gray!20}{Gray} highlights mean our methods outperform the baselines in the group (i.e., DAG-FDD vs. Original/$\text{DRO}_{\chi^2}$, DAW-FDD vs. Original/Naive/FRM/Group DRO/$Cons$. EFPR).}}

\end{table*}

\subsection{Results on DF-Platter Dataset}
\label{appendix:dfplatter}
We apply our methods and baseline to the Xception network on a recent Deepfake dataset with demographic annotations, namely DF-Platter, to further illustrate the effectiveness of our methods. We mainly consider Gender (Male and Female) and Age (Young Adult, Adult, Old) attributes based on the official annotations. In addition to the single attribute fairness, we also consider the combined attributes (Intersection) group, including Male-Young Adult (M-Y), Male-Adult (M-A), Male-Old (M-O), Female-Young Adult (F-Y), Female-Adult (F-A), and Female-Old (F-O). We train and evaluate our methods on a subset of the DF-Platter dataset consisting of real and FSGAN-generated data from Set A with C23 compression, and use Dlib~\cite{king2009dlib} for face extraction and alignment. The cropped faces are resized to $380\times380$ for training and testing. Training/validation/test datasets are divided following the official split,  without identity overlapping. Experiment results shown in Table~\ref{tb:dfplatter} demonstrate that our methods outperform baseline for most metrics. 

\begin{table*}[t]
\renewcommand\arraystretch{1}
\center
\scalebox{0.75}{
\begin{tabular}{c |  c  c  c |  c  c  c |  c  c  c | c  c  c  c}
 \hline
 \multirow{3}{*}{Methods} & \multicolumn{9}{c|}{Fairness Metrics (\%) \textdownarrow } & \multicolumn{4}{c}{Detection Metrics (\%)} \\ 
 \cline{2-14}
& \multicolumn{3}{c|}{Gender} & \multicolumn{3}{c|}{Age} & \multicolumn{3}{c|}{Intersection} & \multicolumn{4}{c}{Overall} \\ 
\cline{2-14}
& $G_{\text{FPR}}$  & $F_{\text{FPR}}$  & $F_{\text{EO}}$  & $G_{\text{FPR}}$  & $F_{\text{FPR}}$  & $F_{\text{EO}}$ & $G_{\text{FPR}}$  &  $F_{\text{FPR}}$  & $F_{\text{EO}}$  & AUC \textuparrow & FPR \textdownarrow & TPR \textuparrow & ACC \textuparrow \\
\hline 
Original &  3.70 & 3.70 & 3.92  & 3.43 & 3.90 & 5.03 &  4.96 & 11.94 & 14.56 & 99.93 & 2.80 &  99.82 & 98.54\\ 
\hline
 \makecell{DAG-FDD (Ours)} & \colorbox{gray!20}{3.05} & \colorbox{gray!20}{3.05} & \colorbox{gray!20}{3.18}  & \colorbox{gray!20}{3.40} & \colorbox{gray!20}{3.29} & \colorbox{gray!20}{4.06} &  \colorbox{gray!20}{4.72} & \colorbox{gray!20}{10.35} & \colorbox{gray!20}{12.03} & \colorbox{gray!20}{\textbf{99.97}} & \colorbox{gray!20}{2.42} &  \colorbox{gray!20}{\textbf{99.91}} & \colorbox{gray!20}{98.77}  \\
\hline 
DAW-FDD (Ours) &  \colorbox{gray!20}{\textbf{1.95}} & \colorbox{gray!20}{\textbf{1.95}} & \colorbox{gray!20}{\textbf{2.13}}  & \colorbox{gray!20}{\textbf{1.97}} & \colorbox{gray!20}{\textbf{2.17}} & \colorbox{gray!20}{\textbf{2.96}} &  \colorbox{gray!20}{\textbf{3.27}} & \colorbox{gray!20}{\textbf{6.81}} & \colorbox{gray!20}{\textbf{8.81}} & \colorbox{gray!20}{\textbf{99.97}} & \colorbox{gray!20}{\textbf{1.75}} &  99.82 & \colorbox{gray!20}{\textbf{99.05}}  \\

 \hline
\end{tabular}
}
\caption{\label{tb:dfplatter} \textit{Comparison results with different fairness solutions using the Xception detector on DF-Platter testing set across Gender, Age, and Intersection groups. The best results are shown in \textbf{Bold}. $\uparrow$ means higher is better and $\downarrow$ means lower is better. \colorbox{gray!20}{Gray} highlights mean our methods outperform the Original baseline.}}

\end{table*}

\subsection{Dataset Details}
\label{appendix:numberoftraining2}
We show the total number of train/val/test samples of each dataset and the attributes included in our experiment in Table~\ref{tb:datasets-details}. Specifically, the number of training samples within each subgroup for four datasets is shown in Table~\ref{tb:numberofsample}. 

\begin{table}[h]
\renewcommand\arraystretch{1}
\center
\scalebox{0.8}{
\begin{tabular}{ c | c | c  }
 \hline
  Dataset & \# Samples & Sensitive Attributes \\
 \hline 
   FF++ & 126,956 & Gender~(Male, Female), Race~(White, Black, Asian, Others)  \\
  \hline
   Celeb-DF & 143,273 & Gender~(Male, Female), Race~(White, Black, Others) \\
   \hline
   DFD & 40,246 & Gender~(Male, Female), Race~(White, Black, Others) \\
   \hline
   DFDC & 117,065 & Gender~(Male, Female), Race~(White, Black, Asian, Others) \\
   \hline
\end{tabular}
}
\vspace{-0.2cm}
\caption{\label{tb:datasets-details} \textit{Sample number and attributes in each dataset.}} 
\vspace{-2mm}
\end{table}

\begin{table*}[!h]
\center
\scalebox{0.8}{
\begin{tabular}{ c | c  c | c  c  c  c | c  c  c  c  c  c  c  c }
 \hline
 \multirow{2}{*}{Datasets} & \multicolumn{2}{c|}{Gender} & \multicolumn{4}{c|}{Race} & \multicolumn{8}{c}{Intersection} \\ 
\cline{2-15}
& M  & F & A & B  & W  & O  & M-A  & M-B & M-W  & M-O  &  F-A  & F-B  & F-W & F-O \\ \hline
 FF++ & 33549 & 42590 & 10488 & 2579 & 56724 & 6348 & 2475 & 1468 & 31281 & 4163 & 8013 & 1111 & 31281 & 2185 \\ 
 \hline
 Celeb-DF & 87344 & 6251 & - & 630 & 86583 & 6382 & - & 600 & 81194 & 5550 & - & 30 & 5389 & 832 \\ 
 \hline
 DFD & 16607 & 7227 & - & 8121 & 11911 & 3802 & - & 6482 & 7784 & 2341 & - & 1639 & 4127 & 1461 \\
 \hline
 DFDC &  37911 & 33567 & 4059 & 18909 & 40257 & 8253 & 2144 & 9603 & 21755 & 4409 & 1915 & 9306 & 18502 & 3844 \\
\hline
\end{tabular}
}
\caption{\label{tb:numberofsample} \textit{Number of training samples of each group in the FF++, Celeb-DF, DFD and DFDC datasets. ``-" means group does not exist in the dataset. }}
\end{table*}

\subsection{Detailed Results}
\label{appendix:details}
Detailed test results of each subgroup on four datasets based on four models are presented in this section. Table~\ref{tb:detailedresults_datasets} provides comprehensive metrics of each subgroup on the four datasets, while Table~\ref{tb:detailedresults_models} displays details of the four models. These findings align with the results reported in Tables~\ref{tb:baselines}, \ref{tb:datasets}, \ref{tb:detectors} and Figures~\ref{fig:baron4datasets}, \ref{fig:subgroupbar} of the submitted manuscript.

\begin{table*}[!h]
\renewcommand\arraystretch{1}
\center
\scalebox{0.72}{
\begin{tabular}{ c | c |c | c  c | c  c  c  c | c  c  c  c  c  c  c  c }
 \hline
 \multirow{2}{*}{Datasets} & \multirow{2}{*}{Methods} & \multirow{2}{*}{Metric~(\%)} & \multicolumn{2}{c|}{Gender} & \multicolumn{4}{c|}{Race} & \multicolumn{8}{c}{Intersection} \\ 
\cline{4-17}
&  & & M  & F & A & B  & W  & O  & M-A  & M-B & M-W  & M-O  &  F-A  & F-B  & F-W & F-O \\ \hline

 \multirow{12}{*}{FF++} & \multirow{4}{*}{Original} & AUC & 92.42 & 93.30 & 89.33 & 94.44& 92.93& 97.01& 88.09& 95.21& 92.47&95.43 & 90.33& 93.42 & 93.53&99.40 \\ 
 & & FPR & 19.86& 23.95& 32.67& 24.29& 20.10& 19.58& 25.63& 21.74& 19.01& 18.79& 36.72& 26.27& 21.08& 20.53\\ 
 &  & TPR & 91.84& 96.80& 94.92& 95.66 & 94.09& 96.07& 89.13& 95.69& 91.70& 93.43& 97.96& 95.63& 96.35& 99.86\\
 &  & ACC & 89.80& 93.01& 89.55& 92.17& 91.57& 93.49& 86.12& 93.02& 89.83& 91.55& 91.38& 91.30& 93.20& 96.20\\
\cline{2-17}
&  \multirow{4}{*}{DAG-FDD~(Ours)} & AUC & 96.59& 97.65& 96.74& 96.76& 97.08& 98.76& 93.20& 99.44& 96.55& 98.34& 98.24& 94.19& 97.60& 99.31\\ 
  &  & FPR & 8.67& 10.30& 13.65& 8.57& 9.21& 5.42& 14.29& 9.78& 8.06& 6.08& 13.29& 7.63& 10.26& 4.64\\ 
  &  & TPR & 91.93& 96.51& 94.62& 95.25& 94.28& 93.63& 88.16& 98.43& 92.06& 91.11& 98.02& 91.88& 96.39& 97.25\\
  &  & ACC & 91.82& 95.26& 93.01& 94.58& 93.66& 93.79& 87.66& 97.18& 92.04& 91.55& 95.86& 91.97& 95.19& 96.91\\
 \cline{2-17}
&  \multirow{4}{*}{DAW-FDD~(Ours)} & AUC & 96.91& 98.05& 96.39& 97.92& 97.54& 98.23& 94.63& 97.81& 97.07& 97.24& 97.35& 98.23& 98.11& 99.22\\ 
&    & FPR & 11.29& 11.61& 12.58& 13.33& 11.18& 10.84& 9.24& 15.22& 11.20& 12.71& 14.49& 11.86& 11.17& 8.61\\ 
 &   & TPR & 93.48& 97.15& 94.40& 96.36& 95.47& 95.77& 88.91& 96.08& 93.92& 93.13& 97.29& 96.67& 96.94& 99.57\\
 &   & ACC & 92.65& 95.55& 93.04& 94.67& 94.29& 94.68& 89.29& 94.35& 93.02& 92.23& 95.05& 94.98& 95.48& 98.10\\
\hline
 \multirow{12}{*}{Celeb-DF} & \multirow{4}{*}{Original} & AUC & 87.83& 98.04& -& 91.47& 97.40& 99.91& -& 91.47& -& -& -& -& 98.00& 100\\ 
 & & FPR & 16.47& 11.55& -& 11.55& 13.31& 10.00& -& 11.55& -& -& -& -& 11.81& 0.00\\ 
 &  & TPR & 79.62& 96.74& -& 79.62& 96.74& 100& -& 79.62& -& -& -& -& 96.74& 100\\
 &  & ACC & 81.90& 95.44& -& 82.88& 94.89& 91.67 & -& 82.88& -& -& -& -& 95.42& 100\\
\cline{2-17}
&  \multirow{4}{*}{DAG-FDD~(Ours)} & AUC & 91.61& 98.53& -& 92.56& 98.28& 99.98& -& 92.56& -& -& -& -& 98.51& 100\\ 
  &  & FPR & 3.84& 1.82& -& 2.54& 2.43& 1.33& -& 2.54& -& -& -& -& 1.86& 0.00\\ 
  &  & TPR & 73.43& 88.18& -& 73.43& 88.16& 100& -& 73.43& -& -& -& -& 88.16& 100\\
  &  & ACC & 86.68& 89.75& -& 82.31& 89.90& 98.89& -& 82.31& -& -& -& -& 89.71& 100\\
 \cline{2-17}
&  \multirow{4}{*}{DAW-FDD~(Ours)} & AUC & 88.72& 98.93& -& 91.52& 98.38& 100& -& 91.52& -& -& -& -& 98.91& 100\\ 
&    & FPR & 4.78& 0.97& -& 3.80& 1.90& 0.33& -& 3.80& -& -& -& -& 0.99& 0.00\\ 
 &   & TPR & 70.22& 85.33& -& 70.22& 85.31& 100& -& 70.22& -& -& -& -& 85.31& 100\\
 &   & ACC & 84.79& 87.49& -& 79.81& 87.67& 99.44& -& 79.81& -& -& -& -& 87.43& 100\\
\hline
 \multirow{12}{*}{DFD} & \multirow{4}{*}{Original} & AUC & 92.41& 93.34& -& 95.27& 92.12& -& -& 94.12& 90.85& -& -& 98.39& 93.10& -\\ 
 & & FPR & 23.44& 26.39& -& 19.48& 26.83& -& -& 19.65& 26.78& -& -& 18.18& 26.86 &-\\ 
 &  & TPR & 94.57& 97.14& -& 96.32& 95.95& -& -& 94.33& 94.41& -& -& 100& 97.25& -\\
 &  & ACC & 88.36& 89.68& -& 88.37& 88.48& -& -& 86.22& 88.86& -& -& 95.48& 88.20&-\\
\cline{2-17}
&  \multirow{4}{*}{DAG-FDD~(Ours)} & AUC & 92.68& 93.93& -& 94.93& 92.89& -& -& 93.64& 92.26& -& -& 98.51& 93.58& -\\ 
  &  & FPR & 26.53& 29.44& -& 23.51& 29.59& -& -& 23.75& 28.98& -& -& 21.59& 29.89&-\\ 
  &  & TPR & 95.26& 97.14& -& 97.11& 96.13& -& -& 95.75& 94.96& -& -& 99.62& 97.13&- \\
  &  & ACC & 87.75& 88.72& -& 86.73& 87.70& -& -& 84.44& 88.69& -& -& 94.35& 86.99&-\\
 \cline{2-17}
&  \multirow{4}{*}{DAW-FDD~(Ours)} & AUC & 92.38& 93.77& -& 94.55& 92.68& -& -& 93.23& 91.93& -& -& 98.47& 93.42&- \\ 
&    & FPR & 27.01& 28.41& -& 25.97& 28.34& -& -& 26.54& 27.43& -& -& 21.59& 28.79&-\\ 
 &   & TPR & 94.97& 96.71& -& 96.84& 95.86& -& -& 95.55& 94.64& -& -& 99.25& 96.90& -\\
 &   & ACC & 87.39& 88.75& -& 85.36& 87.93& -& -& 82.74& 88.86& -& -& 94.07& 87.26&-\\
\hline

 \multirow{12}{*}{DFDC} & \multirow{4}{*}{Original} & AUC & 91.19& 93.41& 79.27& 94.69& 92.24& 89.33& 66.96& 92.61& 92.67& 86.82& 99.77& 95.50& 91.54& 94.58\\ 
 & & FPR & 8.04& 6.40& 9.30& 5.28& 7.72& 8.67& 20.96& 4.98& 6.61& 13.02& 0.80& 5.57& 9.09&0.99\\ 
 &  & TPR & 74.69& 77.41& 56.68& 81.36& 76.45& 68.15& 44.44& 71.80& 77.80& 67.57& 93.33& 85.14& 75.45& 68.61\\
 &  & ACC & 86.90& 86.83& 87.31& 90.66& 85.72& 84.00& 73.36& 90.34& 87.91& 82.22& 98.94& 90.90& 83.53&86.44\\
\cline{2-17}
&  \multirow{4}{*}{DAG-FDD~(Ours)} & AUC & 90.70& 94.22& 82.44& 95.79& 92.00& 89.73& 69.71& 94.49& 91.18& 87.02& 99.63& 96.29& 92.22& 95.60\\ 
  &  & FPR & 7.22& 5.91& 6.81& 3.87& 7.60& 8.47& 15.28& 3.49& 6.54& 12.54& 0.64& 4.24& 8.91&1.28\\ 
  &  & TPR & 71.97& 76.04& 52.50& 80.92& 74.74& 62.38& 42.22& 70.41& 74.85& 64.36& 83.33& 85.06& 74.67& 60.77\\
  &  & ACC & 86.69& 86.54& 89.14& 91.51& 85.09& 82.32& 77.74& 91.25& 86.92& 81.80& 98.63& 91.70& 83.25&83.03\\
 \cline{2-17}
&  \multirow{4}{*}{DAW-FDD~(Ours)} & AUC & 93.30& 96.24& 88.66& 98.23& 93.64& 93.69& 77.89& 96.73& 93.24& 91.43& 100& 98.97& 93.87& 97.72\\ 
&    & FPR & 5.06& 3.34& 4.88& 1.95& 5.43& 4.31& 11.35& 3.04& 4.99& 6.27& 0.16& 0.91& 5.97& 0.85\\ 
 &   & TPR & 74.11& 75.77& 54.17& 82.59& 74.74& 65.15& 40.00& 74.56& 76.92& 65.35& 96.67& 85.76& 73.14& 64.99\\
 &   & ACC & 88.84& 87.93& 91.05& 93.36& 86.36& 86.04& 80.66& 92.45& 88.65& 86.77& 99.70& 94.03& 84.06&85.03\\
\hline
\end{tabular}
}
\caption{\label{tb:detailedresults_datasets} \textit{
Detailed test set results of each group in Xception on the FF++, Celeb-DF, DFD and DFDC datasets. '-' means not applicable.}} 
\end{table*}

\begin{table*}[t]
\renewcommand\arraystretch{1}
\center
\scalebox{0.7}{
\begin{tabular}{ c | c |c | c  c | c  c  c  c | c  c  c  c  c  c  c  c }
 \hline
 \multirow{2}{*}{Models} & \multirow{2}{*}{Methods} & \multirow{2}{*}{Metric~(\%)} & \multicolumn{2}{c|}{Gender} & \multicolumn{4}{c|}{Race} & \multicolumn{8}{c}{Intersection} \\ 
\cline{4-17}
&  & & M  & F & A & B  & W  & O  & M-A  & M-B & M-W  & M-O  &  F-A  & F-B  & F-W & F-O \\ \hline

 \multirow{12}{*}{ResNet-50} & \multirow{4}{*}{Original} & AUC & 93.54& 95.15& 92.19& 96.38& 94.51& 96.10& 87.06 & 97.67 & 94.22 & 94.04 & 94.75 & 96.16 & 94.92 & 98.23 \\ 
 & & FPR & 24.57& 27.15& 33.28& 27.62& 24.97 & 20.48& 35.29 & 20.65 & 23.16 & 24.86 & 32.13 & 33.05 & 26.61 & 15.23 \\ 
 &  & TPR & 94.24& 98.30& 96.89& 97.68& 96.17 & 96.49& 93.65 & 96.67 & 94.17 & 94.04 & 98.59 & 98.75 & 98.06 & 100 \\
 &  & ACC & 90.96 & 93.65& 91.01& 93.25 & 92.42& 93.69 & 87.75 & 94.02 & 91.15 & 91.12 & 92.76 & 92.48 & 93.60 & 97.27 \\
\cline{2-17}
&  \multirow{4}{*}{DAG-FDD~(Ours)} & AUC & 93.50& 95.56& 92.40& 95.21& 94.69& 96.58& 89.36& 97.17& 93.78& 94.82& 93.92& 94.82& 95.70& 98.82\\ 
  &  & FPR & 21.33& 23.54& 27.76& 26.67& 21.32& 21.69& 27.31& 17.39& 20.20& 25.41& 28.02& 33.90& 22.34& 17.22\\ 
  &  & TPR & 93.16& 97.32& 96.48& 95.96& 95.03& 96.01& 93.86& 95.88& 92.83& 93.64& 97.85& 96.04& 97.10& 99.42\\
  &  & ACC & 90.64& 93.51& 91.76& 92.00& 92.12& 93.09& 89.55& 93.85& 90.56& 90.69& 92.94& 90.13& 93.59& 96.43\\
 \cline{2-17}
&  \multirow{4}{*}{DAW-FDD~(Ours)} & AUC & 92.78& 94.78& 91.78& 95.79& 93.84& 95.50& 88.59& 96.75& 93.17& 93.23& 93.29& 95.93& 94.80& 97.81\\ 
&    & FPR & 21.52& 25.31& 29.29& 23.81& 22.55& 22.29& 29.83& 19.57& 19.76& 27.07& 28.99& 27.12& 25.07& 16.56 \\ 
 &   & TPR & 90.29& 96.72& 94.66& 95.66& 93.20& 95.00& 90.74& 94.90& 89.75& 91.62& 96.72& 96.46& 96.46& 99.86\\
 &   & ACC & 88.23& 92.69& 90.00& 92.25& 90.40& 92.15& 86.55& 92.69& 88.09& 88.73& 91.84& 91.81& 92.57& 96.91\\
\hline
 \multirow{12}{*}{EfficientNet-B3} & \multirow{4}{*}{Original} & AUC & 94.72& 97.07& 94.56& 98.80& 95.96& 96.85& 90.78& 98.64& 94.87& 96.37& 96.59& 99.00& 97.03& 98.29\\ 
 & & FPR & 19.19& 21.17& 25.61& 19.05& 19.65& 16.57& 24.79& 18.48& 19.13& 12.71& 26.09& 19.49& 20.11& 21.19\\ 
 &  & TPR & 96.07& 98.25& 97.33& 99.19& 97.19& 96.07& 93.00& 99.02& 96.56& 93.74& 99.60& 99.38& 97.78& 99.42\\
 &  & ACC & 93.42& 94.70& 92.86& 96.00& 94.20& 93.99& 89.37& 96.35& 93.83& 92.74& 94.73& 95.65& 94.55& 95.72\\
\cline{2-17}
&  \multirow{4}{*}{DAG-FDD~(Ours)} & AUC & 97.01& 97.46& 96.06& 99.45& 97.27& 97.73& 94.67& 99.74& 97.08& 97.26& 96.68& 99.19& 97.51& 98.81\\ 
  &  & FPR & 8.14& 8.61& 10.43& 0.95& 8.46& 8.43& 9.66& 0.00& 8.37& 8.29& 10.87& 1.70& 8.55& 8.61\\ 
  &  & TPR & 90.32& 95.20& 93.77& 94.34& 92.50& 94.05& 86.55& 94.71& 90.48& 90.40& 97.57& 93.96& 94.40& 99.28\\
  &  & ACC & 90.59& 94.51& 92.95& 95.17& 92.33& 93.64& 87.32& 95.52& 90.68& 90.61& 95.97& 94.82& 93.87& 97.86\\
 \cline{2-17}
&  \multirow{4}{*}{DAW-FDD~(Ours)} & AUC & 95.96& 96.68& 95.80& 98.09& 96.22& 97.79& 95.09& 97.45& 95.83& 97.20& 95.92& 98.69& 96.68& 98.88\\ 
&    & FPR & 8.24& 8.20& 9.51& 9.05& 8.16& 5.72& 8.82& 11.96& 8.24& 5.53& 9.90& 6.78& 8.09& 5.96\\ 
 &   & TPR & 88.55& 94.05& 93.36 & 95.15& 90.68& 92.98& 86.44& 93.73& 88.34& 89.50& 97.00& 96.67& 92.90& 97.97\\
 &   & ACC & 89.11& 93.64& 92.80& 94.42& 90.89& 93.19& 87.40& 92.86& 88.93& 90.27& 95.69& 95.99& 92.72 &97.27 \\
\hline
 \multirow{12}{*}{DSP-FWA} & \multirow{4}{*}{Original} & AUC & 89.75& 93.97& 90.13& 95.60& 91.72& 94.16& 83.99& 96.38& 90.10& 90.63& 93.32& 96.51& 93.62& 98.77\\ 
 & & FPR & 28.48& 34.37& 40.64& 31.43& 29.58& 34.94& 38.24& 20.65& 26.50& 37.02& 42.03& 39.83& 32.37& 32.45\\ 
 & & TPR & 90.08& 95.99& 95.33& 96.77& 92.46& 94.11& 91.82& 95.29& 89.48& 90.30& 97.17& 98.33& 95.28& 99.57\\
 &  & ACC & 86.85& 90.45& 88.33& 91.83& 88.55& 89.31& 85.69& 92.86& 86.70& 86.08& 89.73& 90.80& 90.29& 93.82\\
\cline{2-17}
&  \multirow{4}{*}{DAG-FDD~(Ours)} & AUC & 90.19& 92.78& 89.50& 95.43& 91.79& 91.83& 84.10& 95.13& 90.78& 91.04& 92.08& 96.06& 92.90& 93.54\\ 
  &  & FPR & 29.86& 34.50& 42.64& 37.14& 30.11& 31.63& 42.86& 41.30& 27.25& 29.83& 42.51& 33.90& 32.71& 33.78\\ 
  &  & TPR & 91.02& 96.15& 94.96& 98.59& 93.14& 93.99& 88.38& 98.63& 90.98& 89.90& 98.42& 98.54& 95.18& 99.86\\
  &  & ACC & 87.39& 90.55& 87.64& 92.33& 89.01& 89.76& 82.01& 92.53& 87.80& 86.85& 90.65& 92.14& 90.15& 93.82\\
 \cline{2-17}
&  \multirow{4}{*}{DAW-FDD~(Ours)} & AUC & 88.15& 93.54& 90.54& 94.44& 90.63& 92.05& 85.08& 95.23& 87.81& 91.13& 94.30& 93.86& 93.49& 94.07\\ 
&    & FPR & 28.81& 31.83& 34.20& 31.43& 29.19& 34.94& 26.05& 38.04& 27.88& 35.91& 38.89& 26.27& 30.37& 33.78\\ 
 &   & TPR & 87.64& 95.92& 92.73& 96.06& 91.22& 95.18& 82.35& 95.88& 87.10& 92.42& 98.19& 96.25& 95.12& 99.13\\
 &   & ACC & 84.77& 90.85& 87.49& 91.25& 87.60& 90.21& 80.63& 90.70& 84.49& 88.04& 91.16& 91.81& 90.52& 93.22\\
\hline

 \multirow{12}{*}{RECCE} & \multirow{4}{*}{Original} & AUC & 97.15& 98.86& 97.44& 98.65& 98.12& 98.51& 94.75& 99.20& 97.31& 97.76& 98.71& 98.40& 98.89& 99.71\\ 
 & & FPR & 21.67& 20.80& 32.67& 28.10& 19.26& 13.86& 39.50& 41.30& 19.07& 11.05& 28.74& 17.80& 19.43& 17.22\\ 
 & & TPR & 97.03& 99.29& 98.04& 100& 98.17& 97.80& 95.37& 100& 97.10& 96.47& 99.43& 100& 99.18& 99.71\\
 & & ACC & 93.77& 95.62& 92.06& 95.08& 95.08& 95.88& 88.26& 93.69& 94.29& 95.30& 94.09& 96.49& 95.82& 96.67\\
\cline{2-17}
&  \multirow{4}{*}{DAG-FDD~(Ours)} & AUC & 97.71& 98.90& 97.13& 99.65& 98.40& 99.04& 94.02& 99.69& 97.97& 98.09& 98.56& 99.62& 98.85& 99.88\\ 
  &  & FPR & 11.71& 12.26& 18.87& 6.19& 11.45& 7.83& 19.75& 4.35& 11.08& 10.50& 18.36& 7.63& 11.80& 4.64\\ 
  &  & TPR & 95.15& 98.31& 96.55& 98.79& 96.79& 96.13& 92.47& 99.22& 95.35& 94.04& 98.70& 98.33& 98.16& 99.13\\
  &  & ACC & 93.95& 96.38& 93.55& 97.92& 95.33& 95.48& 89.97& 98.67& 94.23& 93.34& 95.46& 97.16& 96.36& 98.45\\
 \cline{2-17}
&  \multirow{4}{*}{DAW-FDD~(Ours)} & AUC & 97.73& 98.98& 98.17& 98.85& 98.27& 99.00& 95.52& 98.24& 97.94& 98.31& 99.54& 99.49& 98.70& 99.89\\ 
&    & FPR & 8.29& 8.04& 7.67& 10.00& 8.64& 3.01& 7.56& 16.30& 8.56& 2.76& 7.73& 5.09& 8.72& 3.31\\ 
 &   & TPR & 92.24& 96.74& 93.70& 97.37& 94.60& 94.29& 85.79& 98.24& 92.81& 90.81& 97.85& 96.46& 96.29& 99.28\\
 &   & ACC & 92.15& 95.86& 93.43& 96.08& 94.02& 94.73& 87.15& 96.01& 92.57& 91.80& 96.79& 96.15& 95.38& 98.81\\
\hline
\end{tabular}
}
\caption{\label{tb:detailedresults_models} \textit{
Detailed test set results of each group in ResNet-50, EfficientNet-B3, DSP-FWA, and RECCE on the FF++ dataset.}} 
\end{table*}